\documentclass{article}

\PassOptionsToPackage{sort&compress}{natbib}
\usepackage[dvipsnames,svgnames,x11names]{xcolor}
\usepackage[utf8]{inputenc}
\usepackage[final]{corl_2026} % Uncomment for the camera-ready ``final'' version.
\usepackage{amsthm}
\theoremstyle{plain}

\theoremstyle{definition}

\theoremstyle{remark}

\usepackage[title]{appendix}
\usepackage{multicol}
\usepackage{changes}
\usepackage{inconsolata} % font
\usepackage[font=normal,labelfont=bf]{caption}
\usepackage{subcaption}

\usepackage{dsfont}
\usepackage{bm}

\usepackage{multirow}
\usepackage{amsthm,amssymb,lipsum}
\usepackage{bm}
\usepackage{times}
\usepackage{mathrsfs}
\usepackage{enumitem}
\usepackage{xltabular} % longtable + tabularx
\usepackage{array}
\usepackage{colortbl}
\usepackage{wrapfig2}
\usepackage{bbding}
\usepackage{pifont}
\usepackage{xspace}
\usepackage{amsmath}
\usepackage{wasysym}
\usepackage{textcomp}
\usepackage{microtype}      % microtypography
\usepackage[bottom]{footmisc}
\makeatletter
\def\@fnsymbol#1{\ensuremath{\ifcase#1\or \dagger\or \ddagger\or
  \mathsection\or \mathparagraph\or \|\or **\or \dagger\dagger
  \or \ddagger\ddagger \else\@ctrerr\fi}}
\makeatother
\makeatletter
\providecommand\@makenormalcolbox{%
  \setbox\@outputbox\vbox to\@colht{%
    \@texttop
    \unvbox\@outputbox
    \@textbottom
  }%
}
\providecommand\@makespecialcolbox{%
  \@makenormalcolbox
  {\setbox\@tempboxa\box\@kludgeins}%
}
\makeatother
\usepackage{algpseudocode}
\usepackage{algorithm}

\usepackage{makecell}

\usepackage{color}
\usepackage{tocloft}
\usepackage{caption}
\usepackage{float}
\usepackage{afterpage}

\usepackage{stackengine}

\usepackage{changes}

\usepackage{bbm}

\usepackage{cuted}
\usepackage{duckuments}

\usepackage{tabularx}

\usepackage[percent]{overpic}
\usepackage{adjustbox}
\usepackage{utfsym}

\def\eg{{\it e.g.}\xspace}

\def\ie{{\it i.e.}\xspace}
\def\vs{{\it vs.}\xspace}

\newif\ifHEROCVPRLayout
\HEROCVPRLayoutfalse
\providecommand{\HEROwrapwidth}{0.50\textwidth}

\providecommand{\HEROappendixref}[2]{\cref{#1}}
\newenvironment{HEROwrapfigure}[3][0]{%
  \ifHEROCVPRLayout
    \begin{figure}[t]
    \centering
  \else
    \ifnum#1>0
      \wrapfigure[#1]{#2}{#3}
    \else
      \wrapfigure{#2}{#3}
    \fi
    \centering
  \fi
}{%
  \ifHEROCVPRLayout
    \end{figure}
  \else
    \endwrapfigure
  \fi
}
\newenvironment{HEROwraptable}[3][0]{%
  \ifHEROCVPRLayout
    \begin{table}[t]
    \centering
  \else
    \ifnum#1>0
      \wraptable[#1]{#2}{#3}
    \else
      \wraptable{#2}{#3}
    \fi
    \centering
  \fi
}{%
  \ifHEROCVPRLayout
    \end{table}
  \else
    \endwraptable
  \fi
}

\definecolor{mydarkblue}{rgb}{0,0.08,0.55}
\definecolor{drp-blue}{HTML}{1f77b4}
\definecolor{drp-blue}{HTML}{1f77b4}
\definecolor{pretty-blue}{RGB}{0, 113, 188}
\definecolor{pretty-green}{RGB}{57,181,74} % kaiming green
\definecolor{mypurple}{RGB}{55,0,168} % kaiming green
\definecolor{icmlblue}{rgb}{0,0.08,0.45} % ICML Blue
\definecolor{linecolor1}{HTML}{F1F7FB}
\definecolor{linecolor2}{HTML}{E3EFF7}
\definecolor{linecolor3}{HTML}{D5E4F0}
\definecolor{cvprblue}{rgb}{0.21,0.49,0.74}
\definecolor{myblue}{rgb}{.39,.58,.93}
\definecolor{runpei-orange}{HTML}{F35F27}
\definecolor{runpei-blue}{HTML}{14294B}
\definecolor{datacolor}{HTML}{0009BF}
\definecolor{vitcolor}{HTML}{fc8e62}

\definecolor{curvecolor1}{HTML}{0AB7A9}
\definecolor{curvecolor2}{HTML}{6355CE}
\definecolor{curvecolor3}{HTML}{EA553A}

\newcommand{\pit}{\pi_\text{t}\xspace}

\newcommand{\DeltaEt}{\textcolor{runpei-orange}{\bm{\Delta\mathcal{E}_t}}}

\renewcommand{\HEROwrapwidth}{0.50\textwidth}

\renewcommand{\HEROappendixref}[2]{\cref{#1}}
\newcommand{\HEROAbstractFigure}{}
\providecommand{\SuspendTagging}[1]{}
\providecommand{\ResumeTagging}[1]{}
\usepackage{soul}
\sethighlightmarkup{\IfIsColored{{\sethlcolor{authorcolor!30}\hl{#1}}}{#1}}

\def\ours{{\scshape HERO}\xspace}

\usepackage{graphicx}
\usepackage{booktabs}
\usepackage{caption}
\usepackage{float}
\usepackage{colortbl}
\usepackage{lipsum}
\usepackage{changes}

\IfFileExists{titletoc.sty}{%
  \usepackage{titletoc}%
}{%
  \newcommand{\startcontents}{}%
  \newcommand{\printcontents}[3]{}%
}

\usepackage[capitalize]{cleveref}
\Crefname{section}{Sec.}{Secs.}
\crefname{section}{Sec.}{Secs.}
\Crefname{table}{Tab.}{Tabs.}
\crefname{table}{Tab.}{Tabs.}
\Crefname{figure}{Fig.}{Figs.}
\crefname{figure}{Fig.}{Figs.}
\Crefname{assumption}{Assumption}{Assumptions}
\crefname{assumption}{Assumption}{Assumptions}
\Crefname{theorem}{Theorem}{Theorems}
\crefname{theorem}{Theorem}{Theorems}
\Crefname{proposition}{Proposition}{Propositions}
\crefname{proposition}{Proposition}{Propositions}

\makeatletter
\def\blfootnote{\xdef\@thefnmark{}\@footnotetext}
\makeatother

\hypersetup{
   pdfauthor={Runpei Dong, Ziyan Li, Arjun Gupta, Xialin He, Saurabh Gupta},
   pdftitle={HERO: Learning Humanoid End-Effector Control for Visual Whole-Body Open-Vocabulary Object Grasping},
   pdfsubject={Robot Learning, Robotics, Humanoid Robots},
   pdfkeywords={Humanoid Robots; Visual Loco-Manipulation; Open-Vocabulary Loco-Manipulation; Dexterous Whole-Body Grasping},
   colorlinks=true,
   linkcolor=blue,
   citecolor=YellowGreen,
   urlcolor=blue,
   filecolor=blue
}

\usepackage{nth}

\title{HERO: Learning \underline{H}umanoid \underline{E}nd-Effector Cont\underline{ro}l for Visual Whole-Body Open-Vocabulary Object Grasping}

\author{
Runpei Dong\thanks{Equal contribution.}\qquad
Ziyan Li\footnotemark[1]\qquad
Arjun Gupta\qquad
Xialin He\qquad
Saurabh Gupta\\
\vspace{0.05in}
\textit{University of Illinois Urbana-Champaign}\\
\href{https://hero-humanoid.github.io/}{\textcolor{runpei-orange}{\texttt{\textbf{hero-humanoid.github.io}}}}
}

\begin{document}
% Use plain title marks; footmisc and hyperref otherwise create missing links.
\begingroup
\makeatletter
\def\@footnotemark{\@makefnmark}
\let\H@@footnotemark\@footnotemark
\makeatother
\maketitle
\endgroup
\setcounter{footnote}{0}

\begin{center}
    \centering
    \captionsetup{type=figure}
    \vspace{-8pt}
    \includegraphics[width=\linewidth]{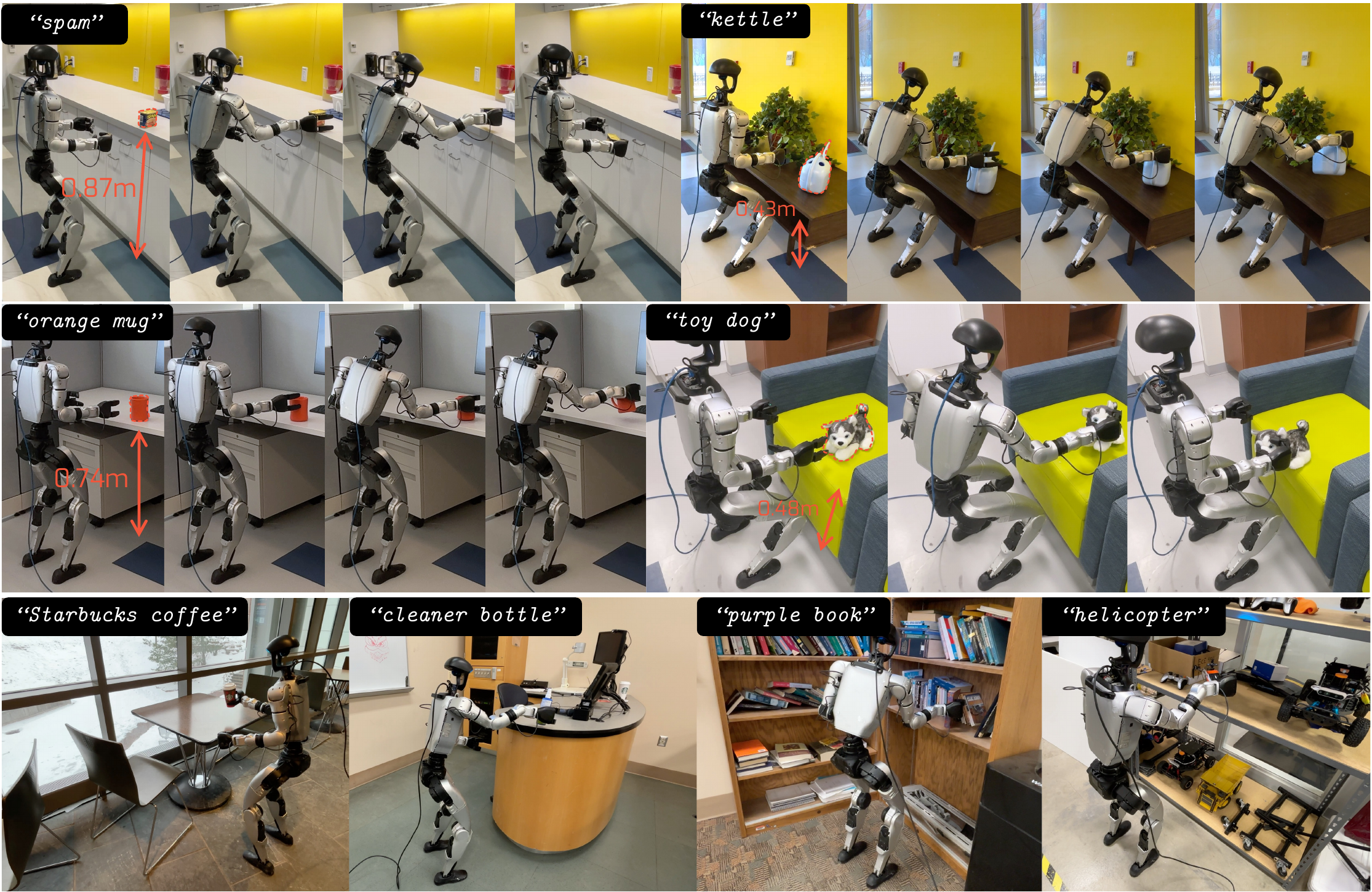}
    \caption{\textbf{Open-vocabulary visual loco-manipulation with humanoids.}
    We enable a humanoid to \textit{autonomously} loco-manipulate \textit{novel} objects in \textit{novel} scenes using onboard sensors. A modular system combines large vision models for visual generalization with an accurate end-effector tracking policy. It achieves an $\mathit{83.8\%}$ average success rate at reaching and picking up novel objects in challenging real-world scenes that require whole-body bending, squatting, and twisting.
    }\label{fig:teaser}
\end{center}

\begin{abstract}
\providecommand{\HEROAbstractFigure}{\begin{figure*}[!t]
  \includegraphics[width=1.0\linewidth]{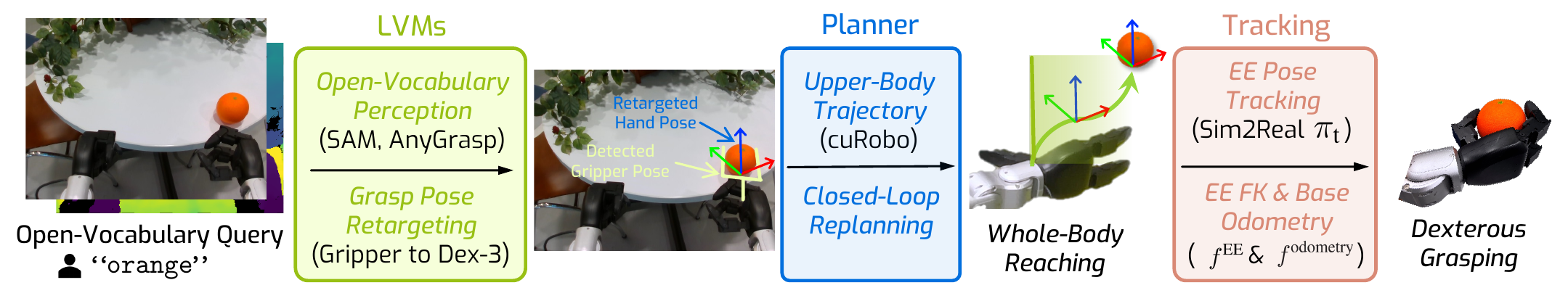}
  \vspace{-12pt}
  \caption{\textbf{Overall architecture for our proposed modular system for open-vocabulary object grasping.} Given a free-form natural language text query indicating which object needs to be picked, we use open-vocabulary large vision models (LVMs: Grounding DINO~\cite{GroundingDINO24} and SAM~\cite{SAM325}) to segment out the object of interest and predict parallel jaw grasps (using the AnyGrasp model~\cite{AnyGrasp23}). We retarget the predicted grasp to a Dex-3 hand. We use our proposed whole-body end-effector tracker to convey the robot arm to the predicted grasp before picking up the object. By decomposing {\it action planning} (\ie identifying which object to pick and using what grasp) from {\it action execution} (\ie actual control of the robot), we inherit the strong visual generalization from pre-trained models as well as strong control capabilities for simulated training of the tracking policy.
  \vspace{-10pt}
  }\label{fig:system}
\end{figure*}}
\HEROAbstractFigure
Visual loco-manipulation of arbitrary in-the-wild objects requires accurate
end-effector (EE) control and a generalizable understanding of the scene from
visual inputs (\eg, RGB-D images). Existing imitation and sim2real methods
jointly learn both these aspects via monolithic end-to-end learning and are
thus hard to scale. In this work, we bring to bear the best tools for each
of these problems -- large vision models for generalizable scene understanding
and simulated training for accurate EE control -- leading to an overall modular
loco-manipulation system that exhibits strong generalization. Our core
technical innovation is HERO, an accurate residual-aware EE tracking policy
made possible by combining classical robotics with machine learning. It uses a)
inverse kinematics to convert residual end-effector targets into reference
trajectories, b) a learned neural forward model for accurate forward
kinematics, and c) goal adjustment and replanning.  Together, these innovations
reduce the end-effector tracking error to 2.44cm, outperforming the strongest
prior method by 5.5$\times$. Our overall system operates in diverse
real-world environments, from offices to coffee shops, where the robot reliably
grasps various everyday objects (\eg, mugs, apples, toys) on surfaces ranging
from 43cm to 92cm in height. Systematic modular and end-to-end tests
demonstrate the effectiveness of our proposed design. We believe our advances
open up new ways of training humanoids to interact with daily objects.

\end{abstract}

% \keywords{Humanoid Robots, Visual Whole-body Open-vocabulary Grasping, End-Effector Control, Generalization}

% ----------------------------------

\section{Introduction}
\vspace{-5pt}
Think about reaching to pick up the objects placed on the various table tops in
\cref{fig:teaser}. As humans we can reliably and robustly use our whole bodies
to execute such pick ups.
We can reach across a table with our back, rotate our torso for objects to the
side, or squat for low coffee tables, all while balancing on two legs.
We can pick up seen objects on seen tables, but also novel objects on novel
tables in novel scenes. Once we have glanced at the object and scene, we can
even do this with our eyes closed. In this paper, we develop a framework that
equips humanoids with this fundamental capability: {\it autonomously reach over to
pick up novel objects in novel everyday environments.}

Humanoids are doing
backflips~\cite{OmniRetarget25,BeyondMimic25,HumanUP25,TaoHuangStand25,ASAP25},
so why would we be writing about such a mundane and seemingly unimpressive
task? There are two key differences that make the problem of manipulating novel
objects harder. First, in-the-wild operation means that neither reference
motions nor privileged sensors (\eg MOCAP) are available. But instead we need
to process high-dimensional RGB-D image observations to infer object locations
and plan trajectories. Second, object manipulation requires {\it precise} and
{\it goal-directed} behavior: a robot needs to get its hand {\it where the
object actually is}, different from a backflip where the focus is on landing
safely rather than at a precise location. Operation in novel environments,
sensing using on-board RGB-D sensors, precise end-effector (EE) control, and maintaining
balance while reaching around, make this problem challenging.

The state-of-the-art for training humanoids for such tasks is end-to-end imitation
learning in the real world~\cite{HumanPlus24, HelixFigureAI25, TeslaAI21,
OmniH2O24, LargeBehaviorModelTRI25} or end-to-end visual RL in
simulation~\cite{VIRAL25}. However, the difficulty of real-world data collection
and of setting up photo- and physically-realistic scenes in
simulation~\cite{VIRAL25} limits the amount of data diversity and thus the generalization
capabilities of learned policies. This causes them to fall
short of the goal of manipulating {\it novel} objects in {\it novel}
environments. In this paper, we pursue an alternate approach.  We take
inspiration from strong results with modular systems for tabletop object
manipulation problems~\cite{ManipGen25,SoFAR25,OKRobot24}. We use large vision
models to translate high-level instructions (\eg \texttt{\small grasp the red
coke can}) into actionable plans by identifying the target objects in complex
scenes and synthesizing a grasp on them; and a simulation-trained low-level
control module then conveys the robot EE to the grasp location. Being able to use
large pre-trained models enables broad generalization and even open-vocabulary
reasoning. In many ways, this is the more direct, obvious, and performant way
to build such a system. So why is such a modular method not already the go-to
approach for building a humanoid object manipulation system?

While it is easy to get a Franka Emika robot to where you want, it turns out it
is current methods aren't able to accurately control a humanoid
hand. SONIC~\cite{SONIC25}, the leading tracker achieves only a 13.38 cm tracking
error in the real world, an error rate that is simply too large for object
manipulation.%
\footnote{Existing works do not focus on end-effector tracking accuracy.
Imitation learning stacks built on top of inaccurate trackers still work
because the IL policy learns to correct for the inaccurate tracking.}
Our key technical contribution is to develop, \ours, an accurate EE tracking policy
that cuts this error down by 5.5$\times$ to 2.44cm.  This unlocks the
possibility of developing modular humanoid systems for object manipulation that
generalize without large-scale real-world imitation demos.

\begin{figure*}[!t]
  \includegraphics[width=1.0\linewidth]{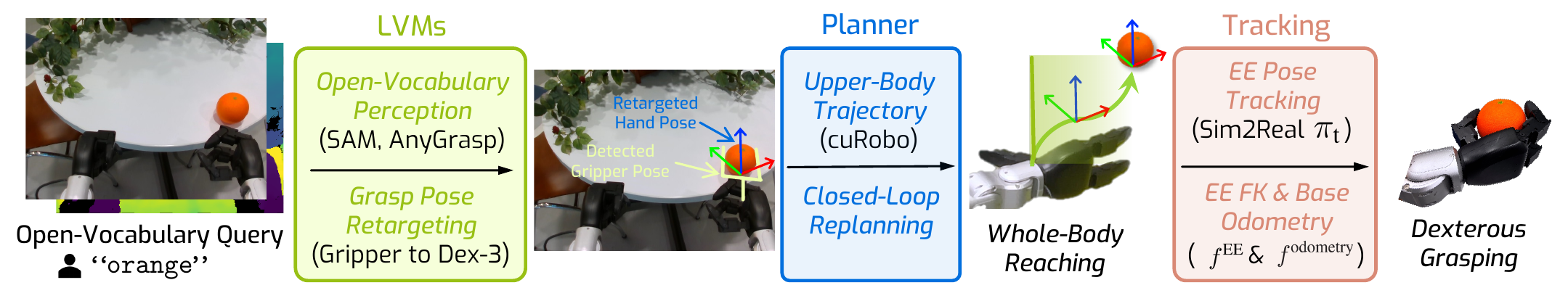}
  \vspace{-12pt}
  \caption{\textbf{Overall architecture for our proposed modular system for open-vocabulary object grasping.} Given a free-form natural language text query indicating which object needs to be picked, we use open-vocabulary large vision models (LVMs: Grounding DINO~\cite{GroundingDINO24} and SAM~\cite{SAM325}) to segment out the object of interest and predict parallel jaw grasps (using the AnyGrasp model~\cite{AnyGrasp23}). We retarget the predicted grasp to a Dex-3 hand. We use our proposed whole-body end-effector tracker to convey the robot arm to the predicted grasp before picking up the object. By decomposing {\it action planning} (\ie identifying which object to pick and using what grasp) from {\it action execution} (\ie actual control of the robot), we inherit the strong visual generalization from pre-trained models as well as strong control capabilities for simulated training of the tracking policy.
  \vspace{-10pt}
  }\label{fig:system}
\end{figure*}

So what are the ingredients of building a highly accurate end-effector tracker?
Our accurate end-effector control algorithm is based on multiple innovations.
First, rather than directly trying to get the end-effector to the target
location, we use a motion planner to generate an upper-body reference motion
that gets the end-effector to the desired target. Second, the policy receives
as input not just the current and target joint angles (output of the motion
planner), but also the current and target end-effector position.  Third, it is
important to obtain a high-quality estimate of the current end-effector
position, as we found that the default analytical forward kinematics and odometry on a
low-cost humanoid robot like Unitree G1 are not accurate. We mitigate this
issue by training neural forward models: an upper body model that provides
accurate EE pose relative to the base, and a base odometry model that provides
an accurate base pose relative to the stationary feet. 
Fourth, because whole-body balancing causes the planned reference to drift relative to the target, we periodically replan from the current state and odometry-corrected goal in a closed loop. 
Finally, a goal-adjustment loop iteratively nudges the commanded pose to remove the systematic sim2real gap.

Using this performant end-effector tracking policy, we develop a modular
system for picking up open-vocabulary novel objects in novel everyday
environments. This modular system leverages an open-vocabulary perception module
to detect and segment the target object using large pre-trained vision
models (Grounding DINO 1.5~\cite{GroundingDINO24} and SAM-3~\cite{SAM325}). 
We next use the AnyGrasp model~\cite{AnyGrasp23} to
produce parallel jaw grasps on the candidate object. We retarget them to the
Dex-3 dexterous hand on the Unitree robot. Finally, we use our tracker as a low-level
controller to achieve the grasp pose. In real-world testing for grasping
open-vocabulary object queries in novel environments, our system achieves a
success rate of 90\% on 10 daily objects across standard and short table
heights, 73.3\% success rate on generalization to 10 daily scenes, and 80\%
success rate on cluttered scenes.

% \vspace{-10pt}
\section{Related Works}
\vspace{-5pt}
\noindent\textbf{Loco-manipulation via Motion Tracking.}
Motion tracking is a central tool for humanoid loco-manipulation: teleoperation or generated references provide whole-body motions, and RL policies learn to track them for sim-to-real transfer~\cite{DeepMimic18,Human2HumanTeleoperation24,OmniH2O24,HumanPlus24,Exbody24,HumanoidGPT26}. Recent work improves tracking accuracy, agility, robustness, reachability, and generalization~\cite{Exbody2_24,Hover24,AMO25,WBEETracking25,ASAP25,GMT25,SONIC25,UnleashingHumanoidReaching25,HiWET26,CEER26,SMASH26}, including force- and stability-aware end-effector control~\cite{HoldMyBeer25,Falcon25,jang2025seec}. Teleoperation systems further enable imitation learning for rich manipulation~\cite{TeleopHumanoidMoCap12,BimanualTeleoperationHumanoid19,Homie25,TWIST25,Opt2Skill25,MobileTelevision24,HMC25,SUGAR26,HumanoidMimicGen26}. In contrast, \ours does not require human teleoperation at test time; it converts visual grasp goals into end-effector targets and executes them with a learned tracker.

\noindent\textbf{Visual Loco-Manipulation.}
Visual loco-manipulation often uses imitation data collected by teleoperation~\cite{HelixFigureAI25,TeslaAI21,HumanPlus24,OmniH2O24,UMIOnLegs25} or reference-state policies that map depth observations to high-level commands or generated motions~\cite{VisualWBControlLocoManip24,VisualMimic25,ActiveSpatialBrain26,VOFA26}. End-to-end visual RL has also shown strong results on selected object categories~\cite{VIRAL25}, and recent foundation-model approaches aim at universal humanoid policies~\cite{PsiZero26,UniT26}. These systems remain limited by collected demonstrations, generated assets, or object categories. \ours instead pairs open-vocabulary perception with modular task-space control, enabling novel language queries and objects without task-specific teleoperation demonstrations.

\noindent\textbf{System Identification.}
Real robots deviate from their nominal models because of hardware inaccuracies, joint elasticity, and dynamics~\cite{ElasticityHumanoid07,HumanoidElasticCalibration21,CalibrationElasticHumanoid22}. System identification mitigates this sim-to-real gap either through online adaptation~\cite{PolicyTransferStrategy19,LeanringFastAdapt20,RMA21,DeepWholeBodyControl22,AtheleticLocoManipulation25,CTS24} or offline correction from collected data~\cite{ASAP25,DBLP:conf/rss/TanZCIBHBV18,DBLP:journals/scirobotics/LeeDBTKH19}. Our learned residual FK and base-odometry models follow the offline direction, using MOCAP data to correct the task-space signals required by the tracking policy.

\section{\textcolor{runpei-orange}{\textbf{HERO}}: \textcolor{runpei-orange}{\textbf{H}}umanoid \textcolor{runpei-orange}{\textbf{E}}nd-Effector Cont\textcolor{runpei-orange}{\textbf{RO}}l}
\label{sec:method}
\vspace{-5pt}
\begin{HEROwrapfigure}[15]{r}{\HEROwrapwidth}
  \vspace{-3pt}
  \includegraphics[width=1.0\linewidth]{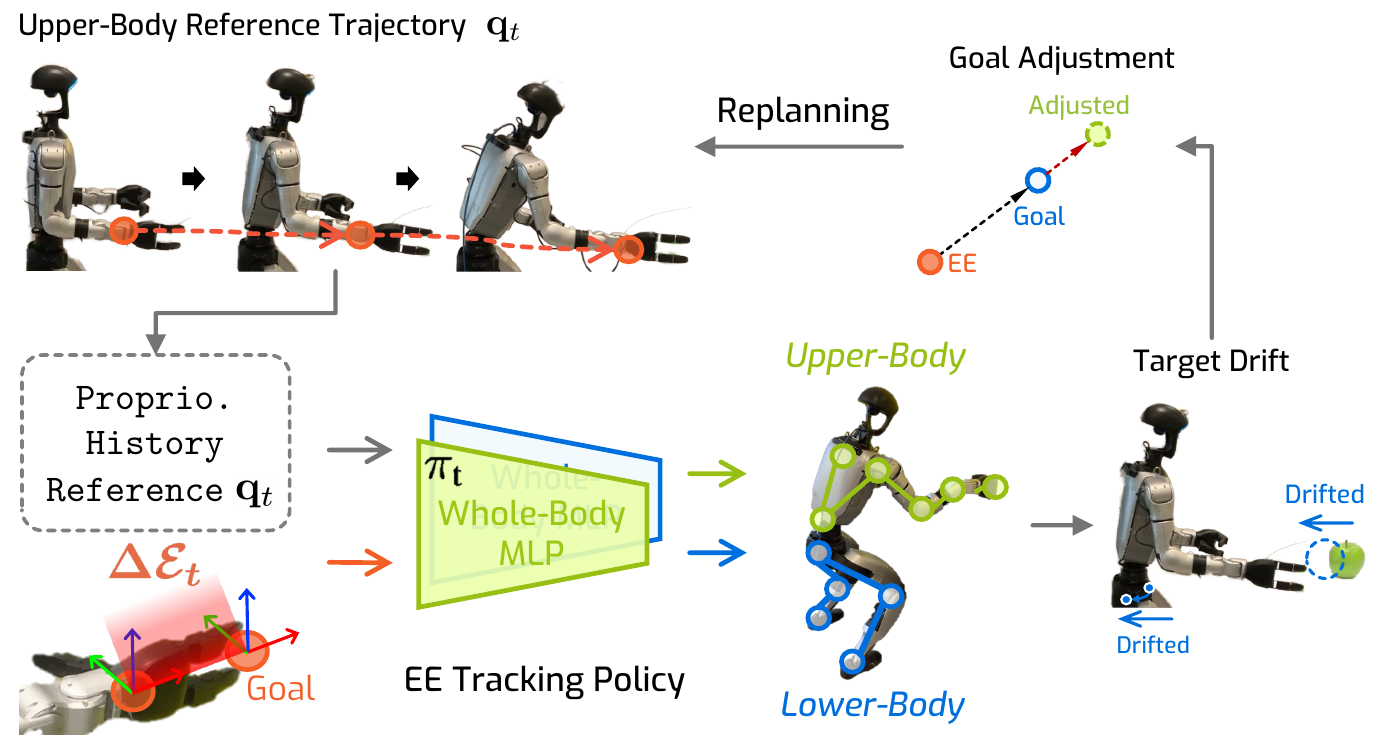}
  \caption{\textbf{\ours end-effector control.} HERO outputs IK and planned references from an EE goal, then tracks them with a learned whole-body policy, which leverages residual EE error computed from learned FK and odometry. Replanning and goal adjustment compensate drift and sim2real bias.
  }\label{fig:policy}
\end{HEROwrapfigure}

Given a desired end-effector pose in the robot frame, \ours outputs motor
commands for all 29 DoFs of a Unitree G1 humanoid~\cite{UnitreeG124}. Although
the feet remain planted, far-reaching motions require coordinated waist, torso,
and leg motion. We therefore combine classical IK and motion planning with
learned task-space feedback instead of learning a monolithic pose-to-action
mapping. The overall tracker is shown in \cref{fig:policy}. 
\ours first converts the target $T^{\text{EE}}\in\text{SE}(3)$ into a base
height $h$ and upper-body joint goal ${\bf q}^*\in\mathbb{R}^{17}$ using IK,
then uses a collision-free planner~\cite{Curobo23} to produce reference joint
and end-effector trajectories. A learned whole-body policy $\pit$ tracks these
references with joint position commands at 50Hz. To make the residual
task-space feedback accurate, $\pit$ uses learned end-effector FK $\eta$ and
base odometry $\xi$. During execution, we also replan periodically and apply a
small goal adjustment to compensate systematic sim-to-real bias.

\subsection{Whole-body End-Effector Tracking Policy, $\pit$}
\label{sec:pit}
\vspace{-5pt}
To track the target EE pose $T^\text{EE}\in \text{SE}(3)$ defined in the
robot frame, our whole-body EE tracking policy $\pit$ first obtains an
upper-body reference trajectory $\{{\bf{q}}_t\}_{t=1}^T$ and the corresponding
6-DoF EE reference trajectory $\{{ee}_t\}_{t=1}^T$ from a motion planner.
Given the trajectory, the current proprioceptive state ${\bf s}_t$, and other
commands, $\pit$ outputs the 29-DoF joint angles commands that are passed
to per-joint PD controllers. 

\textbf{Residual-Aware End-Effector Tracking.}
$\pit$ output actions ${\bf a}_t$ at time $t$ as follows: 
\begin{equation}
{\bf a}_t=\pit\left({\bf s}_t, h_t, {\bf q}_t, \DeltaEt, \mathbf{v}_t, {\bf s}_{t-5:t-1}, {\bf a}_{t-5:t-1}\right),
\end{equation} 
where ${\bf s}_t$ is the current proprioception, $h_t$ is the reference base
height, ${\bf q}_t$ are the reference upper-body joint angles, $\mathbf{v}_t$ are the linear and angular velocity locomotion commands, and ${\bf s}_{t-5:t-1},
{\bf a}_{t-5:t-1}$ are five time steps of proprioception and action history.
The proprioception include the robot's joint angles, joint velocities, angular
velocity, projected gravity, and roll and pitch encoded from
the IMU. We don't use the IMU yaw as it is inaccurate~\cite{OmniH2O24}.
$\DeltaEt$ represents the residual pose error between the current and target end-effector pose in the robot frame, \ie,
\begin{equation}
\DeltaEt
= f^\text{EE}({\bf x}_t) \ominus {ee_t},
\label{eq:se3_residual}
\end{equation}
where $f^\text{EE}({\bf x}_t)$ maps the arm states ${\bf x}_t\in\mathbb{R}^{17}$ to the end-effector pose $T_t^\text{EE}\in \text{SE}(3)$, and $\ominus$ is the inverse pose composition operator.\footnote{We use $\oplus$ to denote pose composition: ${\bf T}_1 \oplus {\bf T}_2 = {\bf T}_1 \cdot {\bf T}_2$ and $\ominus$ for inverse composition: ${\bf T}_1 \ominus {\bf T}_2 = {\bf T}_2^{-1} \cdot {\bf T}_1$.}

\textbf{Architecture and Training.}
$\pi_t$ uses two three-hidden-layer MLPs that share the same observation but separately predict upper- and lower-body actions; their outputs are combined into 29-DoF joint commands. We train $\pit$ in simulation with PPO~\cite{PPO17}, using AMASS~\cite{AMASS19} ($\sim$8K motion sequences) and a curated set of everyday reaching targets ($\sim$8K). Targets are sampled in the robot frame from $[0.1\text{m}, -0.5\text{m}, 0.65\text{m}]$ to $[0.5\text{m}, 0.5\text{m}, 1.15\text{m}]$, with yaw in $[-60^\circ, 60^\circ]$. A motion planner converts each target into upper-body and end-effector reference trajectories for policy training.

\WFclear
\begin{figure*}[!t]
  \includegraphics[width=1.0\linewidth]{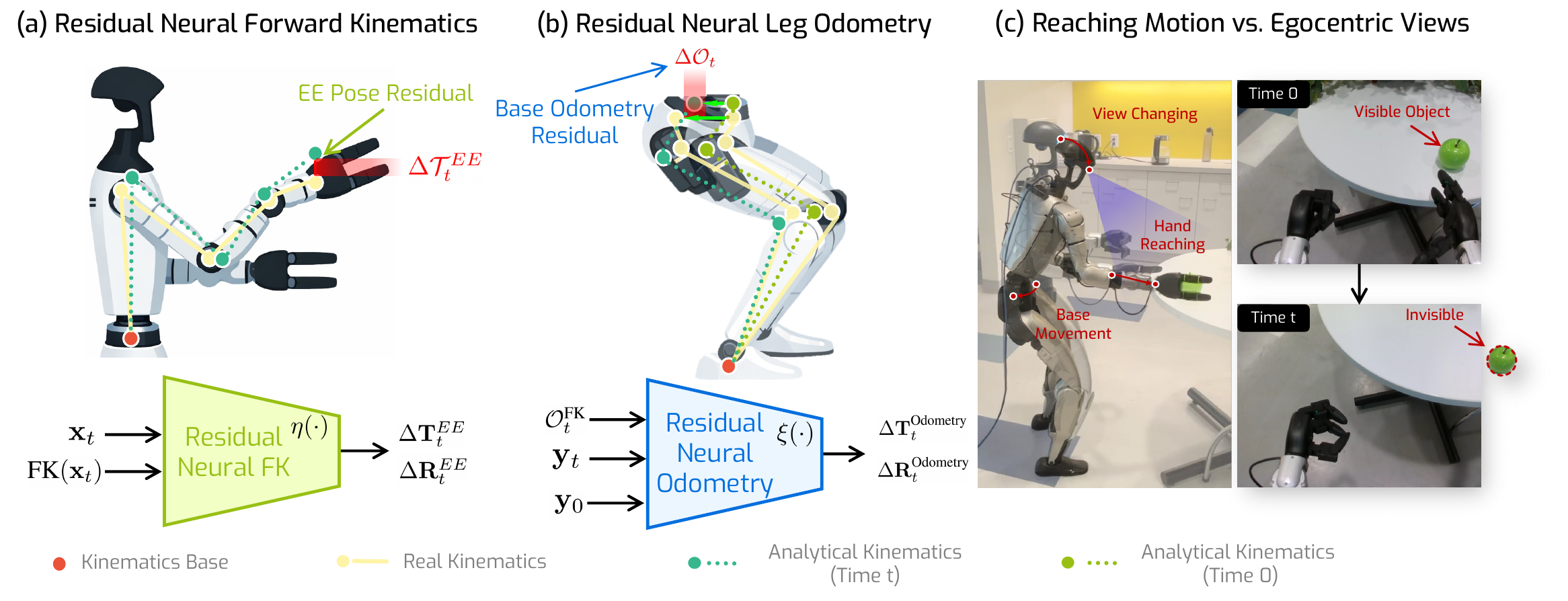}
  \caption{\textbf{Learned neural forward kinematics and odometry models.}
    (a) A residual FK model $\eta$ corrects analytical FK by predicting EE translation and rotation residuals.
    (b) A residual odometry model $\xi$ estimates base motion from lower-body joints.
    (c) Whole-body reaching can move the object out of the egocentric view, making purely visual correction unreliable.
  }\label{fig:neural_fk}
\end{figure*}

\subsection{Learned Residual Neural Forward Models}
\vspace{-5pt}
\textbf{Residual Neural FK.}
\label{sec:forward}
Our residual neural forward kinematics function, $\eta$, learns a correction to the output of the analytical forward kinematic function, $\text{FK}$, to output accurate end-effector poses. Specifically, given the current proprioceptive state of one operating arm
and waist ${\bf x}_t \in \mathbb{R}^{10}$ at timestep $t$, and output from analytical FK, $\text{FK}({\bf x}_t)$, 
the final end-effector pose $f^{\text{EE}}({\bf x}_t)$ is obtained
as:
\begin{equation} f^{\text{EE}}({\bf x}_t) = \text{FK}({\bf x}_t) \oplus \eta\big({\bf x}_t, \text{FK}({\bf x}_t)\big). \end{equation}
Note that the analytical forward kinematics function $\text{FK}({\bf x}_t)$ uses the robot geometry and coordinate transformations to compute the 6-DoF end-effector pose in the robot base frame. For precise robots, FK is itself quite accurate, however as our experiments will show, FK is inaccurate for our humanoid~\cite{ElasticityHumanoid07,HumanoidElasticCalibration21}, necessitating the need for learning a correction. 

\textbf{Residual Neural Leg Odometry.}
\label{sec:neural-odometry}
Different from fixed-base object manipulation tasks, where the robot's base is fixed and stable, humanoid robots' base needs to be dynamically balanced during whole-body reaching.
This movement makes the original reaching target inaccurate, as the reaching target defined in the robot frame is no longer the same place where the object lies.
One might consider using the egocentric visual information for replanning or motion adjustment. 
However, as shown in \cref{fig:neural_fk}, the egocentric view of the robot is too narrow for the robot to see the object when the robot's arm and waist movements are large.
As a result, using the robot's odometry to adjust the reaching goal is critical.
We assume the feet to be static on the ground and use the lower body joint angles to predict the base pose.
By assuming the robot ankle joint as the root joint and the robot base as the end-effector, we can compute the base pose using forward kinematics.

However, similar to the error in analytical FK for EE, analytical FK to compute the base pose is also inaccurate (see analysis in \cref{sec:experiments}). Similar to our solution for EE, we adopt a residual model to mitigate these inaccuracies. We estimate the base pose transformation relative to time step $0$, rather than the absolute base pose, which avoids depending on a global frame and directly tells the controller how much the base has shifted since planning.

Concretely, let ${\bf y}_t \in\mathbb{R}^{6}$ be the 6DOF state of the left (or
right) leg motors. We can get analytical base odometry, \ie base pose relative
to base pose at time step $0$, $\mathcal{O}^{\text{FK}}({\bf y}_t, {\bf y}_0)
\in \text{SE}(3)$, using analytical forward kinematics and SE(3) difference:
$\mathcal{O}^{\text{FK}}({\bf y}_t, {\bf y}_0) = \text{FK}({\bf y}_0) \ominus \text{FK}({\bf y}_t).$
Our residual neural leg odometry model $\xi$ learns the residual:

\begin{equation}
f^{\text{odometry}}({\bf y_t}, {\bf y_0}) = \mathcal{O}^{\text{FK}}({\bf y}_t, {\bf y}_0) \oplus \xi({\bf y}_t, {\bf y}_0, \mathcal{O}^{\text{FK}}({\bf y}_t, {\bf y}_0)). 
\end{equation}

\textbf{Architecture and Training.}
Both residual models are lightweight 3-layer MLPs. $\eta$ predicts a residual
SE(3) correction for the end-effector, while $\xi$ predicts the residual
between MOCAP odometry and analytical leg-FK odometry. Each model uses separate
heads for translation and rotation, outputting a residual translation in
$\mathbb{R}^3$ and a residual rotation via the 6D rotation representation~\cite{ContinuityRotation19}.

We train both models on MOCAP data~\cite{OptitrackSystem}. For $\eta$, a tracking policy sweeps the EE through the workspace while we record motor encoders and MOCAP EE/base poses; marker poses are converted with the Kabsch-Umeyama algorithm~\cite{kabsch1976solution,KULeastSquare02} ($<1.5\text{mm}$ RMSE). We collect 3 hours of data and use the first 2 hours for training and the last 1 hour for validation. $\xi$ uses the same split, with supervision formed by sampling temporal pairs from trajectories. Both models are optimized with MSE losses on residual translations and rotations, using MOCAP-derived $\Delta T^\text{EE}_t$ and $\Delta\mathcal{O}_{n-m}$ as ground truth.

\vspace{-5pt}
\subsection{Replanning and Goal Adjustment}
\label{sec:replan}
\label{sec:adjustment}
\vspace{-8pt}
During execution, the robot drifts from the planned reference because of base-pose changes from whole-body balancing and residual sim-to-real bias, leaving the policy with an out-of-distribution input or a target that is no longer reachable. We therefore replan the remaining reference $\{{\bf q}_t,ee_t\}$ every $k=300$ steps (6\,sec) with cuRobo~\cite{Curobo23} ($\sim$20\,ms), using the current state and the odometry-corrected goal so each fresh trajectory accounts for the latest base displacement.

To further compensate the steady-state EE tracking offset, we maintain a commanded grasp $g$ initialized to $T^{\text{EE}}$ and apply the damped translational update $g^p\gets g^p-\beta,e_t^p$, where $g^p,e_t^p\in\mathbb{R}^3$ denote the translational components of $g$ and the SE(3) residual $e_t = f^{\text{EE}}({\bf x}_t)\ominus T^{\text{EE}}$, $\beta=0.6$, only when $|e_t^p|\le 15\mathrm{cm}$, clipped to $5\mathrm{mm}$/step, and stopped once $|e_t^p|<1.5\mathrm{cm}$. We adjust translation only because scaling the rotation residual the same way does not empirically improve tracking.

\section{A Modular System for Open-Vocabulary Humanoid Object Grasping} 
\label{sec:modular}
\vspace{-10pt}
\begin{figure*}[!t]
  \centering
  \includegraphics[width=1.0\linewidth]{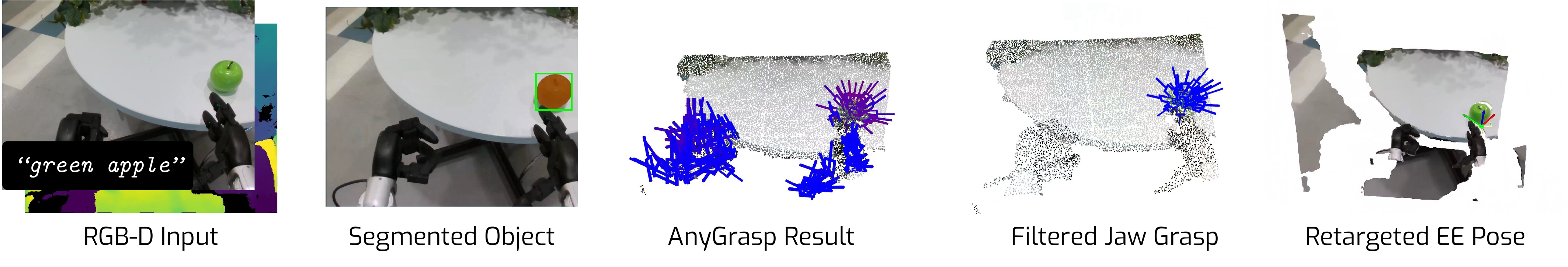}
  \vspace{-5pt}
  \caption{\textbf{Visual perception for language-conditioned grasping.}
  An example of target segmentation, grasp proposals, filtering, and retargeted EE poses from onboard RGB-D observations.
  }\label{fig:visual_perception}
  \vspace{-10pt}
\end{figure*}

\begin{figure}[!b]
  \includegraphics[width=1.0\linewidth]{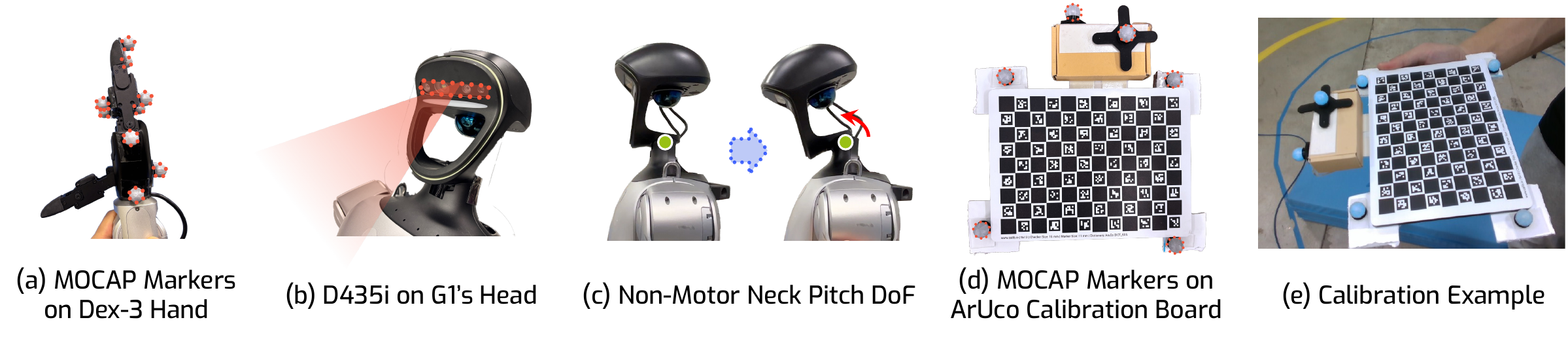}
  \caption{\textbf{MOCAP and camera calibration.}
    (a) Using measured marker-to-link offsets, Kabsch--Umeyama~\cite{kabsch1976solution,KULeastSquare02} recovers 6-DoF EE/base poses from MOCAP coordinates ($<1.5$\,mm RMSE).
    (b--c) The D435i camera is mounted on the G1 head with a passive neck pitch joint.
    (d--e) We manually move an ArUco board~\cite{AruCo14} with MOCAP markers through 60--70 poses, recording board poses in the camera frame and board/base poses in the MOCAP frame. Tsai--Lenz~\cite{TsaiLenz89} then solves the camera-to-base transform. More details can be found in \cref{app:mocap_details} and \cref{app:camera_detail}.
  }\label{fig:mocap_camera2}
\end{figure}

Our task is to pick up novel objects in novel environments from free-form
language queries, using only onboard sensing. Our modular pipeline
builds on \ours: Grounding DINO~\cite{GroundingDINO24} detects the queried
object, SAM-3~\cite{SAM325} segments it, and AnyGrasp~\cite{AnyGrasp23}
proposes parallel-jaw grasps. We filter and retarget the selected grasp
to the Dex-3 hand before \ours executes the reach.
\cref{fig:visual_perception} visualizes the pipeline; detailed filtering and
retargeting steps are given in \cref{app:perception}.
We use MOCAP to accurately calibrate the egocentric camera, \cref{fig:mocap_camera2} shows the MOCAP setup and camera calibration.

For Dex3 retargeting, we rotate the AnyGrasp pose by 45° around the z-axis so
the thumb opposes the other two fingers, improving contact area and force
closure. We also clip end-effector rotation within 70° to avoid twisted IK
postures that degrade tracking.
\WFclear

\section{Experiments}\label{sec:experiments}
\begin{HEROwrapfigure}[9]{r}{\HEROwrapwidth}
    \vspace{-3pt}
  \includegraphics[width=1.0\linewidth]{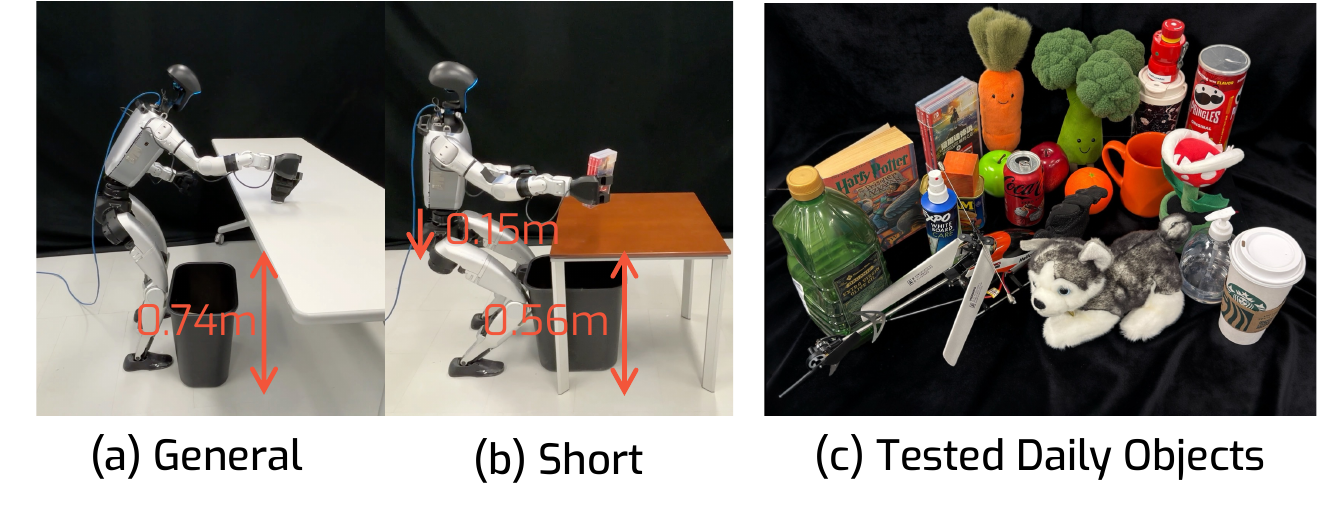}
  \vspace{-20pt}
  \caption{{\bf Novel test setups and objects.}
    (a-b) Standard ($0.74\text{m}$) and short ($0.56\text{m}$) table setups.
    (c) Diverse daily objects used for the test.
  }\label{fig:test_objects}
\end{HEROwrapfigure}

\vspace{-10pt}
We evaluate \ours on open-vocabulary grasping, learned FK, and tracking gains from residual feedback, replanning, and goal adjustment.

Experiments use an {\it unmodified} Unitree G1 humanoid with Dex-3 dexterous hands (3-finger hands), a
head-mounted Intel D435i RGB-D camera, proprioception, and a base IMU. Visual
trials use novel objects and environments; FK and tracking evaluations use a
13-camera Optitrack MOCAP room (see
\HEROappendixref{app:mocap_details}{the supplementary material}). Selected
design choices are validated in Isaac Gym~\cite{IsaacGym21} and
MuJoCo~\cite{Mujoco12}. 
In addition, \cref{app:additional-exp} reports language sensitivity, moving-object grasping, door opening, walking FoV analysis, tracking-error CDFs, trajectory curves, and workspace coverage.

\begin{figure}[t]
\centering
\begin{minipage}[t]{0.51\linewidth}
\vspace{0pt}
\centering
\captionof{table}{{\bf Open-vocabulary grasping success.} Compared with direct PD control and automated SONIC~\cite{SONIC25}, \ours achieves 90\% success across objects at two table heights.}
\label{tab:stand-and-grasp}
\setlength{\tabcolsep}{2pt}
\resizebox{\linewidth}{!}{
\scriptsize
\begin{tabular}{l *{2}{c} *{2}{c} *{2}{c}}
\toprule
& \multicolumn{2}{c}{\bf PD Controller} 
& \multicolumn{2}{c}{\bf SONIC~\cite{SONIC25}} 
& \multicolumn{2}{c}{\bf Ours} \\
\cmidrule(lr){2-3} \cmidrule(lr){4-5} \cmidrule(lr){6-7}
\multirow{2}{*}{\bf Language Query}
  & Gen. & Short 
  & Gen. & Short 
  & Gen. & Short \\
  & 0.74\,m & 0.56\,m 
  & 0.74\,m & 0.56\,m 
  & 0.74\,m & 0.56\,m \\
\midrule
\texttt{\tiny red coke can}          & 2/3 & 0/3 & 0/3 & 2/3 & 3/3 & 3/3 \\
\texttt{\tiny e-stop button}         & 0/3 & 0/3 & 2/3 & 0/3 & 3/3 & 3/3 \\
\texttt{\tiny red piranha plant}     & 0/3 & 0/3 & 0/3 & 0/3 & 3/3 & 3/3 \\
\texttt{\tiny orange cube}           & 0/3 & 0/3 & 0/3 & 0/3 & 3/3 & 3/3 \\
\texttt{\tiny olive oil bottle}      & 0/3 & 1/3 & 0/3 & 0/3 & 2/3 & 2/3 \\
\texttt{\tiny game cartridge}        & 1/3 & 0/3 & 0/3 & 0/3 & 2/3 & 3/3 \\
\texttt{\tiny chip can}              & 1/3 & 1/3 & 0/3 & 0/3 & 2/3 & 3/3 \\
\texttt{\tiny hand soap bottle}      & 0/3 & 0/3 & 2/3 & 3/3 & 3/3 & 3/3 \\
\texttt{\tiny robot hand}            & 0/3 & 0/3 & 0/3 & 0/3 & 3/3 & 2/3 \\
\texttt{\tiny red apple}             & 0/3 & 0/3 & 0/3 & 1/3 & 3/3 & 2/3 \\
\midrule
\bf Total & 4/30 & 2/30 & 4/30 & 6/30 & \bf 27/30 & \bf 27/30 \\
\bottomrule
\end{tabular}}
\end{minipage}
\hfill
\begin{minipage}[t]{0.47\linewidth}
\vspace{0pt}
\centering
\includegraphics[width=\linewidth]{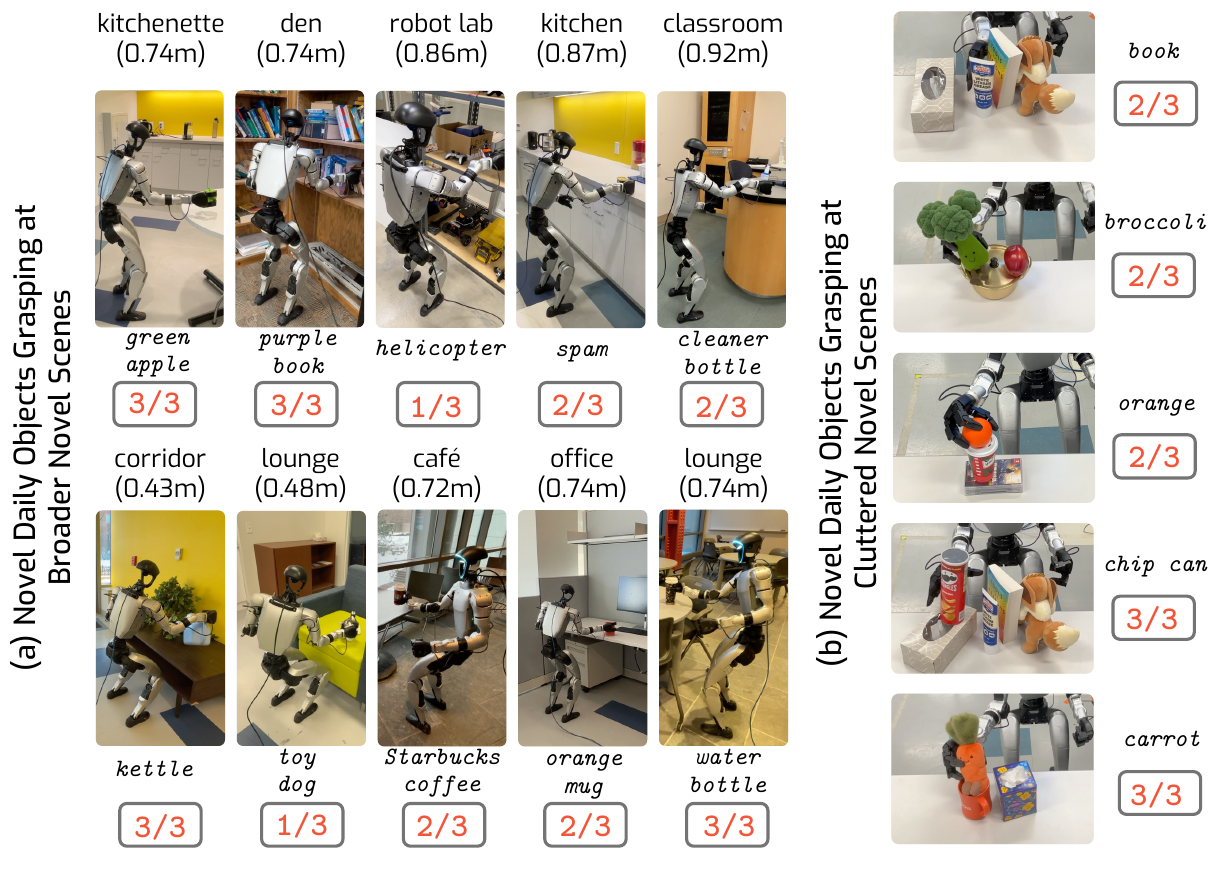}
\caption{\textbf{Open-vocabulary grasping in real-world scenes.}
\ours achieves 22/30 (73.3\%) success across broader novel scenes and 12/15 (80\%) success in cluttered layouts.
}\label{fig:benchmark_generalization}
\end{minipage}
\vspace{-6pt}
\end{figure}

\subsection{End-to-end System Testing}
\vspace{-5pt}
\textbf{Experimental Setup.} Each trial starts 10-20 cm from a table (43-92cm
high). The robot must grasp the queried object using onboard sensing and lift
it for more than 2 seconds. We run 3 trials per object per table height;
\cref{fig:test_objects} shows the objects and \cref{tab:stand-and-grasp} lists
the queries.

\textbf{Baselines.}
We compare \ours to two alternate controllers. To maximally isolate the impact
of the tracking quality, we plug in these alternate controllers into the same
overall modular pipeline as \ours.
\textit{PD Controller} tracks IK reference joints without a learned policy.
For \textit{SONIC}~\cite{SONIC25}, we use the three-point VR teleoperation
interface. We use the predicted grasp as the right EE target 
and place the head
and the left EE targets via FK from upper-body joint positions obtained from right EE IK.

\textbf{Results:} 
\ding{182} \textbf{{10 Daily Objects}.}
\cref{tab:stand-and-grasp} shows that \ours succeeds in 90\% of trials across
objects, queries, and two table heights, \vs 10\% for PD control and
16.7\% for automated SONIC. Accurate whole-body end-effector tracking is
essential for turning visual grasp goals into physical grasps. 
\ding{183} \textbf{{10 Daily Scenes}.} \cref{fig:benchmark_generalization}(a)
evaluates broader scenes such as robot labs and classrooms, where \ours reaches
73.3\% success.
\ding{184} \textbf{{5 Cluttered Layouts}.} 
In cluttered layouts
(\cref{fig:benchmark_generalization}(b)), \ours achieves 80\% success without
teleoperation demonstrations, remaining language-sensitive under distractors.
On 1,450 queries constructed from OCID-Ref images~\cite{OCIDRef21}, our perception module detects targets in 73.8\% of canonical-name queries and 89.7\% of queries with attributes and spatial relations (\cref{tab:ocid-ref}).

\subsection{End-effector Tracking Accuracy Evaluation and Ablation Study}
\vspace{-5pt}
We evaluate tracking on 180 reaching goals in simulation and real MOCAP:
60 poses across three table heights, sampled 5-15cm above the table.
We report EE translation/orientation error against MOCAP and upper-body joint error from motors. Baselines include the PD controller, AMO~\cite{AMO25},
FALCON~\cite{Falcon25}, and automated SONIC~\cite{SONIC25}.
We also conduct ablation study on forward model accuracy (\cref{sec:forward}), closed-loop adapation strategies (\cref{sec:adjustment}).

\begin{table}[t]
\centering
\begin{minipage}[t]{0.48\linewidth}
\centering
\renewcommand{\arraystretch}{1.05}
\setlength{\tabcolsep}{3pt}
\caption{\textbf{Simulation tracking comparison.} End-effector translation, orientation, and joint-tracking errors averaged over three table heights in MuJoCo. \ours attains the lowest error; SONIC lacks in EE tracking feedback and joint position only baselines fall behind.}
\label{tab:tracking_results}
\small
\resizebox{\linewidth}{!}{%
\begin{tabular}{lccc}
\toprule[0.95pt]
\multirow{2}{*}{\bf Method} & \bf Trans. & \bf Orient. & \bf Joint \\
 & \bf (cm) & \bf (deg) & \bf (rad) \\
\midrule[0.6pt]
FALCON~\cite{Falcon25} & $13.57 \pm 4.41$ & $19.33 \pm 5.30$ & $\mathbf{0.02 \pm 0.00}$ \\
AMO~\cite{AMO25}       & $8.29 \pm 3.82$  & $13.85 \pm 5.91$ & $\mathbf{0.02 \pm 0.00}$ \\
SONIC~\cite{SONIC25}   & $4.10 \pm 2.12$  & $16.65 \pm 9.27$ & -- \\
\rowcolor{linecolor1}
\ours (ours)           & $\mathbf{2.48 \pm 1.15}$ & $\mathbf{11.23 \pm 3.97}$ & $0.17 \pm 0.03$ \\
\bottomrule[0.95pt]
\end{tabular}}
\end{minipage}
\hfill
\begin{minipage}[t]{0.48\linewidth}
\centering
\renewcommand{\arraystretch}{1.05}
\setlength{\tabcolsep}{3pt}
\caption{\textbf{Real-world tracking comparison.} MoCap-measured EE errors on hardware. \ours retains sub-3\,cm accuracy in the real world, while
SONIC baseline degrade by 4--5$\times$ compared to simulation.
Lacking dynamic gravity compensation, the PD controller leaves the EE persistently low and misoriented.
}
\label{tab:tracking-real}
\small
\resizebox{\linewidth}{!}{%
\begin{tabular}{lccc}
\toprule[0.95pt]
\multirow{2}{*}{\bf Method} & \bf Trans. & \bf Orient. & \bf Joint \\
 & \bf (cm) & \bf (deg) & \bf (rad) \\
\midrule[0.6pt]
PD Controller         & $12.09 \pm 3.14$ & $40.82 \pm 36.27$ & $0.24 \pm 0.03$ \\
SONIC~\cite{SONIC25}  & $13.38 \pm 1.43$ & $16.75 \pm 10.06$ & -- \\
\rowcolor{linecolor1}
\ours (full)          & $\mathbf{2.44 \pm 0.86}$ & $\mathbf{8.22 \pm 3.52}$ & $\mathbf{0.21 \pm 0.05}$ \\
\bottomrule[0.95pt]
\end{tabular}}
\end{minipage}

\end{table}

\noindent\textbf{Comparisons against state-of-the-art.}
\cref{tab:tracking_results} reports simulation results averaged over three table
heights. \ours achieves the lowest errors, with 2.48cm translation
error versus 4.10cm for SONIC, 8.29cm for AMO, and 13.57cm
for FALCON. SONIC is the closest translation baseline since it
also reasons about EE targets, but its lack of end-effector error
feedback leads to higher tracking error. The gap between joint and task-space
metrics also shows that joint accuracy alone does not ensure end-effector
accuracy. \cref{tab:tracking-real} shows the same trend on real-world MOCAP:
full \ours achieves 2.44cm translation and 8.22$^\circ$ orientation error,
outperforming both PD control (12.09cm, 40.82$^\circ$) and automated SONIC
(13.38cm, 16.75$^\circ$).
In simulation, analytical FK and odometry are exact, so HERO uses them directly; replanning and goal adjustment remain enabled as in real-world deployment without further specifications.

\begin{table}[t]
\centering
\begin{minipage}[t]{0.48\linewidth}
\centering
\renewcommand{\arraystretch}{1.05}
\setlength{\tabcolsep}{3pt}
\caption{\textbf{Forward-model ablation.} Replacing analytic FK with our learned models reduces sim2real bias and approaches the MoCap oracle.
}
\label{tab:tracking-learned-model}
\small
\resizebox{\linewidth}{!}{%
\begin{tabular}{ccccc}
\toprule[0.95pt]
\bf EE & \bf Base & \bf Trans. & \bf Orient. & \bf Joint \\
\bf Pose & \bf Pose & \bf (cm) & \bf (deg) & \bf (rad) \\
\midrule[0.6pt]
FK   & FK   & $4.67 \pm 1.30$ & $14.59 \pm 3.99$ & $0.20 \pm 0.03$ \\
Ours & FK   & $3.35 \pm 0.70$ & $14.07 \pm 3.93$ & $0.19 \pm 0.03$ \\
FK   & Ours & $3.89 \pm 1.06$ & $14.28 \pm 4.75$ & $0.20 \pm 0.04$ \\
\rowcolor{linecolor1}
Ours & Ours & $\mathbf{2.56 \pm 1.23}$ & $\mathbf{12.06 \pm 4.38}$ & $\mathbf{0.18 \pm 0.03}$ \\
\midrule[0.6pt]
\it MoCap & \it MoCap & $\mathit{2.44 \pm 0.86}$ & $\mathit{8.22 \pm 3.52}$ & $\mathit{0.21 \pm 0.05}$ \\
\bottomrule[0.95pt]
\end{tabular}}
\end{minipage}
\hfill
\begin{minipage}[t]{0.48\linewidth}
\centering
\renewcommand{\arraystretch}{1.05}
\setlength{\tabcolsep}{3pt}
\caption{\textbf{Closed-loop feedback ablation.} Real-world MoCap tracking with replanning or goal adjustment removed. Replanning significantly improves tracking accuracy, while goal adjustment further reduces errors.}
\label{tab:tracking-replan}
\small
\resizebox{\linewidth}{!}{%
\begin{tabular}{lccc}
\toprule[0.95pt]
\multirow{2}{*}{\bf Method} & \bf Trans. & \bf Orient. & \bf Joint \\
 & \bf (cm) & \bf (deg) & \bf (rad) \\
\midrule[0.6pt]
\rowcolor{linecolor1}
\ours (full)   & $\mathbf{2.44 \pm 0.86}$ & $\mathbf{8.22 \pm 3.52}$ & $0.21 \pm 0.05$ \\
w/o Replan     & $5.17 \pm 2.21$ & $16.13 \pm 4.66$ & $0.21 \pm 0.03$ \\
w/o Goal Adj.  & $2.71 \pm 0.87$ & $9.38 \pm 2.72$ & $\mathbf{0.20 \pm 0.03}$ \\
\bottomrule[0.95pt]
\end{tabular}}
\end{minipage}

\end{table}

\noindent\textbf{Forward models and closed-loop adaptation.}
\cref{tab:tracking-learned-model} shows that learned EE/base estimates reduce
analytical-FK bias and give the best non-oracle tracking accuracy. Using learned
estimates for both EE and base poses reduces translation error from 4.67cm to
2.56cm, close to the MOCAP oracle at 2.44cm. \cref{tab:tracking-replan} ablates
the feedback terms on real hardware. Removing replanning roughly doubles
translation error from 2.44cm to 5.17cm, while removing goal adjustment gives
a smaller degradation. These results
indicate that learned pose estimation and closed-loop replanning improve accuracy.

\newcommand{\largetext}[1]{{\bf #1}}
\newcolumntype{g}{>{\columncolor{gray!20}}c}

\begin{HEROwraptable}[9]{r}{0.72\linewidth}
\renewcommand{\arraystretch}{1.1}
\centering
\caption{\textbf{FK and odometry accuracy.} EE/base translation and rotation errors from MOCAP; kinematic calibration and residuals reduce bias.}
\label{tab:forward-kinematics}
\vspace{-5pt}
\resizebox{\linewidth}{!}{
\begin{tabular}{l ggcc}
\toprule
\bf \bf \multirow{2}{*}{\bf Method} & \multicolumn{2}{g}{\bf a) End-effector
Pose} & \multicolumn{2}{c}{\bf b) Base Odometry} \\ \cmidrule(lr){2-3}
\cmidrule(lr){4-5}
                                    & \bf Translation & \bf Rotation  & \bf Translation & \bf Rotation  \\
                                    & \bf Error (cm) & \bf Error (deg) & \bf Error (cm) & \bf Error (deg) \\
\midrule
Analytical FK                                    & 1.30     & 6.03     & 1.00     & 0.49     \\
Kinematic Calibration   & 0.58     & \bf 2.83 & \bf 0.51 & 0.51     \\
Learned FK, no residual (ours)                   & 4.44     & 7.06     & 0.53     & 0.75     \\
Learned FK (ours)                                & \bf 0.50 & 3.18     & 0.52     & \bf 0.46 \\
\bottomrule
\end{tabular}
}
\end{HEROwraptable}

\noindent\textbf{Vanilla FK \vs Learned FK.}
\label{sec:forward-kinematics-eval}
We measure FK error by marking the robot base
and EE and using their MOCAP relative transform as ground truth.
We record poses and joint angles, then compare our method to
a) analytical FK and
b)
kinematic calibration, where we learn URDF geometric model parameters~\cite{RobotCalibration87,KinematicConversion22,MUKCa26}.
\cref{tab:forward-kinematics} shows systematic bias in analytical FK for both EE pose and base odometry. FK corrections, through kinematic calibration or our residual model, reduce this bias. Ours achieves lower errors on 2 of 4 metrics, possibly because the neural model may be able to capture non-geometric effects (\eg, elasticity~\cite{ElasticityHumanoid07,CalibrationElasticHumanoid22}). While we use the neural model for HERO, this experiment suggests that the calibrated model would also offer improvements over the default calibration.

\begin{figure}[t]
% \vspace{-8pt}
  \centering
  \includegraphics[trim=0 83 14 0, clip, width=0.49\linewidth]{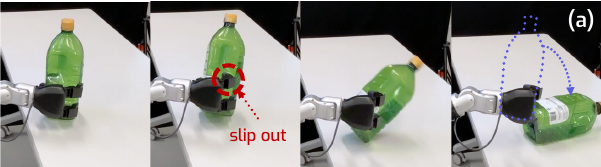}
  \includegraphics[trim=0 0 14 83, clip, width=0.49\linewidth]{figures/src/failure.pdf}
  \caption{\textbf{Failure mode examples}. 
  (a) Object slips from the hand.
  (b) Object knocked over.
  }
  \label{fig:failure_mode}
  % \vspace{-15pt}
\end{figure}

% \vspace{-3pt}
\subsection{Failure Mode Analysis}
% \vspace{-8pt}
\Cref{fig:failure_mode} shows two common failures: \textit{object slipping} on
large or irregular objects, and \textit{object knock-over} when grasp
orientation or hand size contacts unstable objects such as standing books.
% \vspace{-7pt}

\subsection{Extending HERO to Other Tasks Like Door Opening}\label{app:door}
\begin{figure*}[t!]
  \includegraphics[width=1.0\linewidth]{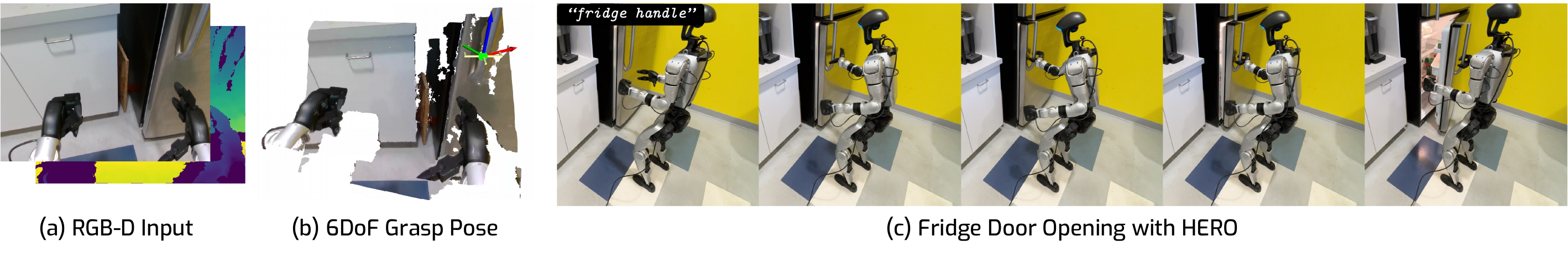}
  \caption{\textbf{Door opening with HERO}. 
  (a) The egocentric RGB-D visual inputs.
  (b) Given a language query for the door handle (\eg, \texttt{``fridge handle''} prompted here), our modular system obtains the grasping pose, same as the pipeline for picking up objects.
  (c) The robot executes the reaching trajectory and closes the hand when reaching the target poses closely, and then returns to the default pose with the door opened.
  Note that while the door is heavy, HERO successfully manages to open the door with a smooth and stable door motion.
  }
  \label{fig:door_opening}
\end{figure*}

As \ours constructed a modular system that coordinates high-level planning and low-level end-effector control, it reveals a possibility of extending \ours to broader tasks like door opening, which is a challenging loco-manipulation task~\cite{OpenDoorHumanoid08,DARPAChallenge18,DoorMan25}.
In \cref{fig:door_opening}, we directly employ HERO to identify the target grasping pose for \texttt{fridge door handle}, followed by the same pipeline as object grasping, enabling the humanoid to grasp the fridge handle and finally open it when returning to the default position.
Note that the door requires a large force to open because of magnetic attraction; we leave the door unlatched beforehand.
This result shows \textit{HERO}'s modular potential, and it is also possible to extend our system to broader loco-manipulation tasks by incorporating off-the-shelf trajectory generation frameworks \cite{CoPa24, ReKep24, SoFAR25, DoorArjunRSS25}.

\section{Discussions and Limitations}
\vspace{-8pt}
We present a modular humanoid system that separates open-vocabulary \textit{action planning} from simulator-trained \textit{action controller}, achieving 90\% success on daily-object grasping without large-scale real-world imitation. 
This decomposition lets large vision models handle visual generalization while a learned motion tracker executes whole-body reaching. The experiments show two lessons: task-space state must be accurate (analytical FK has 1.30cm EE error while ours reaches 0.50cm), and explicit error feedback during closed-loop execution effectively reduces real-world drifts (our tracker reaches 2.5cm error). 
Remaining limitations include restricted egocentric FoV for far or high targets, motivating active neck control and visual servoing~\cite{qi2025coordinated,TWIST2,HMC25}; inefficient motions, motivating learned motion priors; and modular failures~\cite{OKRobot24, DoorArjunRSS25} plus dexterity limits, motivating better hands and retargeting.
\section*{Acknowledgement}
We are grateful to the Coordinated Science Laboratory and the Center for Autonomy for access
to experimental space and the G1 humanoid robot used in this work. We thank John Hart for his
invaluable help with the humanoid robots. This material is based upon work supported by an NSF
CAREER Award (IIS-2143873).
% ----------------------------------
% The CoRL style automatically selects corlabbrvnat.
\bibliography{main}

% ----------------------------------
\clearpage

\begin{center}
\textbf{\textsc{\Large Appendix}}
\end{center}

\newcommand\DoToC{%
    \hypersetup{linkcolor=black}
  \startcontents
  \printcontents{}{1}{\textbf{}\vskip3pt\hrule\vskip3pt}
  \vskip7pt\hrule\vskip3pt
}

\begin{appendices}
\renewcommand{\thesubsection}{\thesection.\arabic{subsection}}

% Change section and subsection title 
\renewcommand{\thesection}{\Alph{section}} % Changes section numbering to A, B, C, etc.
\renewcommand{\sectionname}{}

\renewcommand{\thesubsection}{\Alph{section}.\arabic{subsection}}

\vspace{5pt}
\DoToC
\vspace{10pt}

\hypersetup{linkcolor=blue}

% Contents begin here:
% ---------------------------------------------------------
% Additional Results
% ---------------------------------------------------------
\section{Additional Experimental Analysis}\label{app:additional-exp}
\subsection{Language Sensitivity}\label{app:language}
We evaluate target detection on OCID-Ref images~\cite{OCIDRef21} using 1,450 language queries; \cref{tab:ocid-ref} reports accuracy for selected query types.
\begin{table}[H]
\caption{\textbf{OCID-Ref query sensitivity.} Detection accuracy on our 1,450-query evaluation (58 classes, five views, five query types). Selected types are shown; overall includes all five.}
\label{tab:ocid-ref}
\centering
\small
\begin{tabular}{lr}
\toprule
Query type & Detection accuracy (\%) \\
\midrule
Canonical name & 73.8 \\
Canonical name + attribute & 86.6 \\
Canonical name + attribute + spatial relation & 89.7 \\
\midrule
Overall (all five query types) & 82.6 \\
\bottomrule
\end{tabular}
\end{table}

\begin{figure}[H]
  \includegraphics[width=1.0\linewidth]{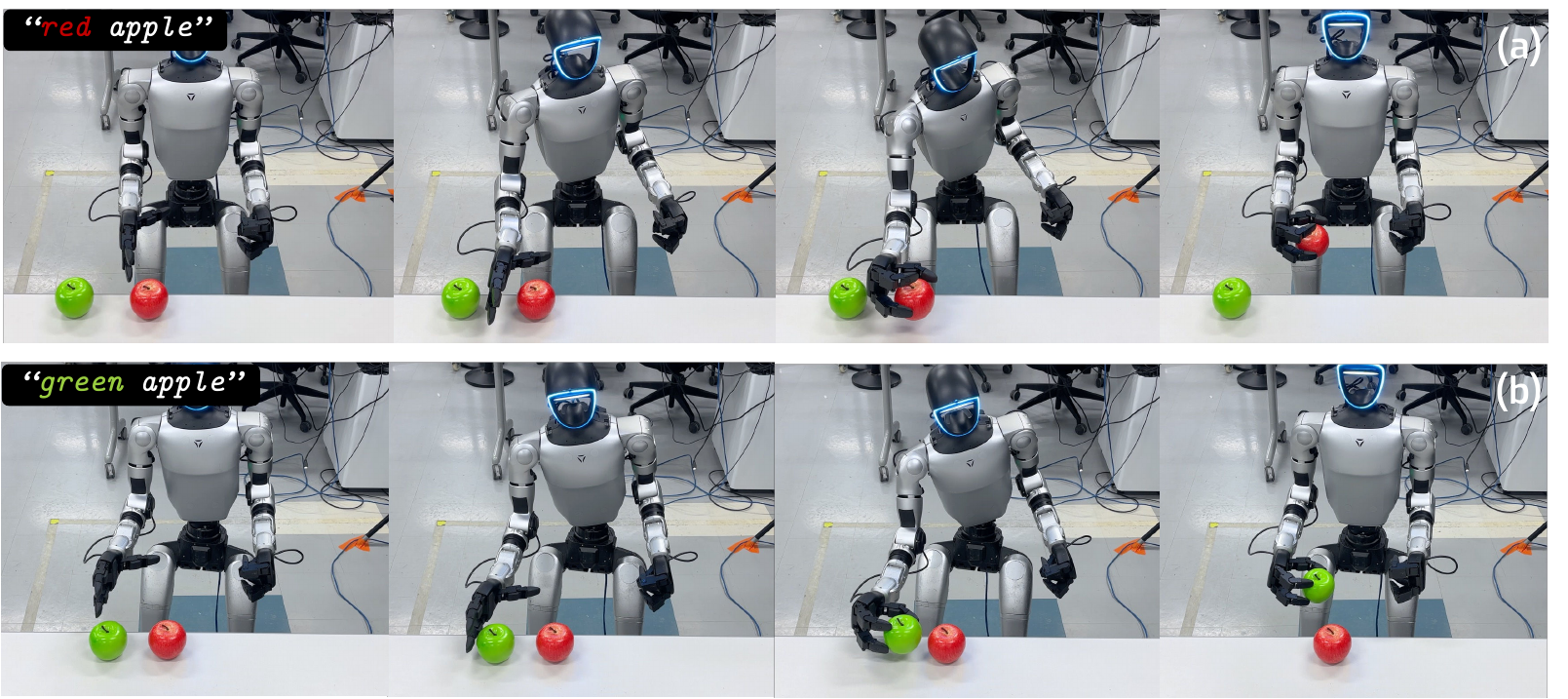}
  \caption{\textbf{HERO is able to distinguish the target object via language queries.}
    (a) Picking up a \texttt{red apple};
    (b) Picking up a \texttt{green apple} instead of \texttt{red apple}.
  }\label{fig:apple_language_sensitivity3}
\end{figure}

\cref{fig:apple_language_sensitivity3} illustrates red- and green-apple selection among distractors in robot grasping.

\subsection{Tracking Error Distribution Analysis}\label{app:cdf}
\begin{figure}[!h]
  \includegraphics[width=1.0\linewidth]{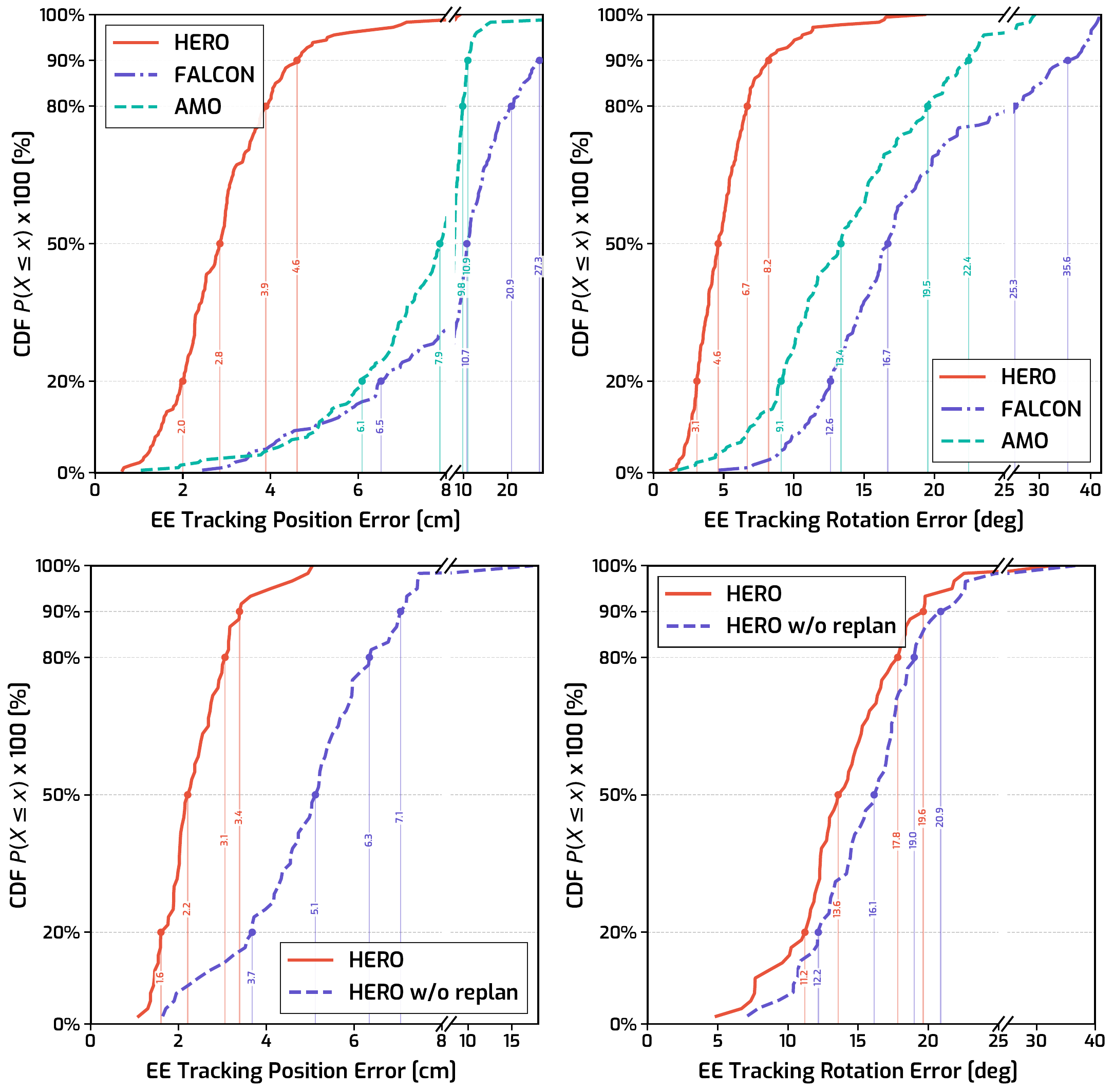}
\caption{\textbf{CDF analysis of end-effector tracking errors.} 
Top row: Comparison of translation (left) and rotation (right) error 
distributions across \ours, FALCON~\cite{Falcon25}, and AMO~\cite{AMO25} for 
all table heights. The steeper curves of \ours indicate consistently lower 
errors and tighter distributions. 
Bottom row: Ablation study showing translation (left) and rotation 
(right) error distributions with and without replanning (\cref{sec:replan}). 
The steeper CDF curves with replanning demonstrate their significant contribution 
to tracking accuracy.}
\label{fig:cdf_analysis}
\end{figure}

We visualize the CDFs of EE tracking errors in \cref{fig:cdf_analysis}.

\noindent\textbf{Top row.}
HERO dominates all baselines: at 80\%, HERO achieves 3.9~cm / 6.7$^\circ$ (pos/rot), versus 20.9~cm / 25.3$^\circ$ for FALCON and 9.8~cm / 19.5$^\circ$ for AMO. At 90\%, HERO remains below 4.6~cm and 8.2$^\circ$, indicating strong tail robustness.

\noindent\textbf{Bottom row.}
Replanning (\cref{sec:replan}) is critical: at 80\%, \ours achieves 3.1~cm vs.\ 6.3~cm without replanning (2.1$\times$ worse), and at 90\% 3.4~cm vs.\ 7.1~cm. Rotation gains are smaller but consistent (median 13.6$^\circ$ vs.\ 16.1$^\circ$).

\begin{figure*}[!t]
  \includegraphics[width=1.0\linewidth]{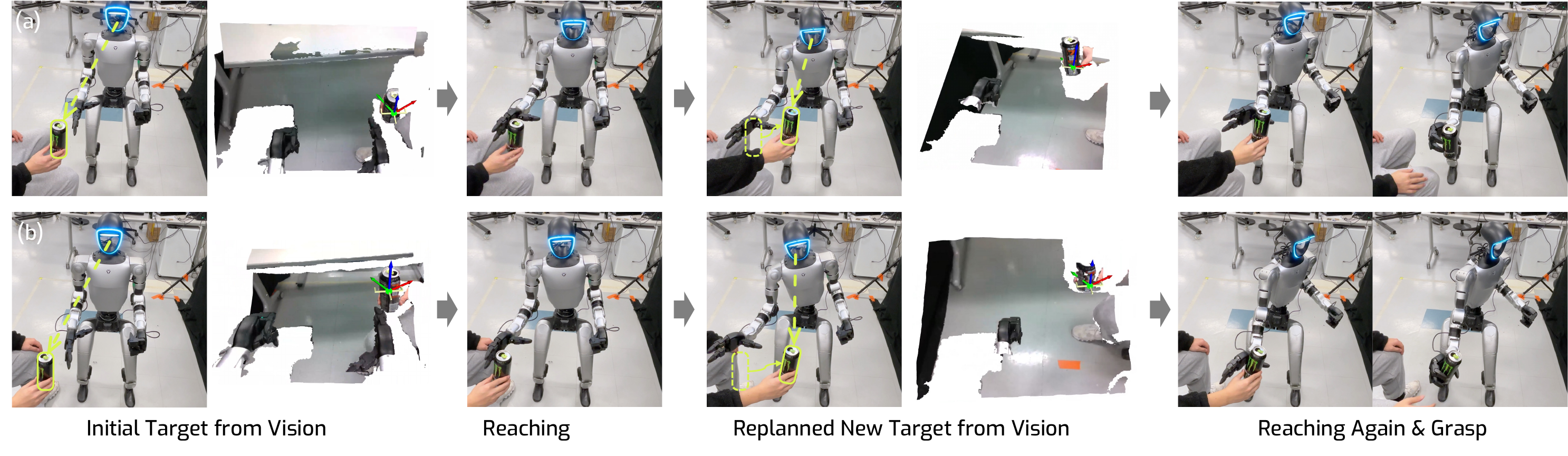}
  \caption{\textbf{HERO enables a humanoid to grasp a moving object via visual closed-loop replanning.}
  (a-b) Two examples of visual closed-loop replanning.
  The goal is to grasp \texttt{black can}.
  }\label{fig:visual_replan}
\end{figure*}
\subsection{Moving Object Grasping with Visual Replanning}\label{app:moving-object}
HERO derives target grasping poses from vision and language queries, enabling the robot to capture moving objects through closed-loop replanning. 
\cref{fig:visual_replan}(a-b) illustrates this dynamic adaptation: while the system initially generates an EE trajectory based on the first vision perception, it re-estimates the pose as the object moves. 
This visual feedback triggers the update of the target grasp, allowing the robot to seamlessly adjust its trajectory and successfully secure the moving object.
Note that, in these two trials, the robot successfully sees the object after moving, but the object can only be seen at the corner due to the rather limited field of view.

\begin{figure}[!t]
  \includegraphics[width=1.0\linewidth]{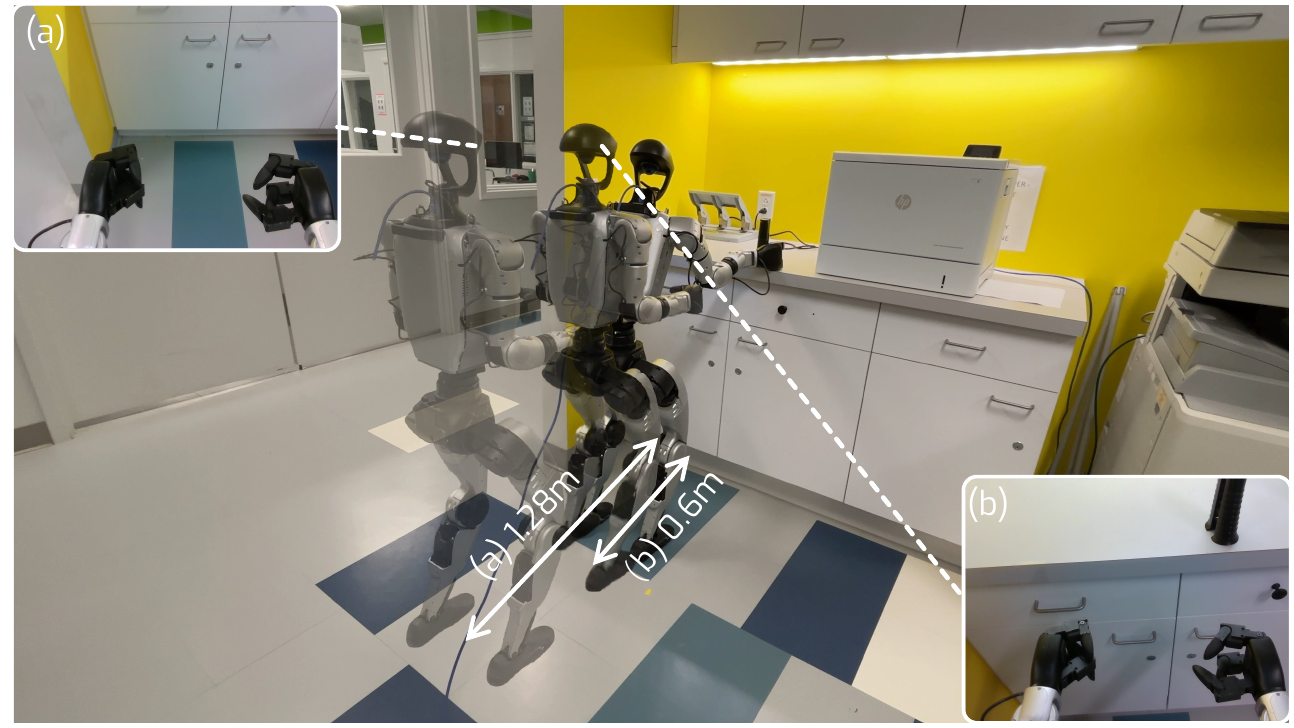}
  \caption{\textbf{Field of view (FoV) visualization using HERO}.
  We let the robot stand at a random distance from the object (\eg, 1.28m), and the robot keeps walking forward until the target object (\texttt{stapler}) is detected.
  After successfully detecting the target object, the robot stops walking and grasps the object via whole-body coordination.
  The robot can only see the object within 0.6m, which makes it hard for the robot to search for the object in a random room.
  } 
  \label{fig:walking_view}
\end{figure}

\subsection{Field of View Analysis}\label{app:fov}
Our system uses the onboard camera for visual perception, which, however, has a limited field of view.
As shown in \cref{fig:walking_view}, the robot first stands randomly at a distance of 1.28m from the object, while the target object (\texttt{stapler}) is not visible at this distance; after walking forward under a consistent velocity command to a distance of 0.6m and continuously detecting the object, the egocentric view successfully captures the target object, which makes the robot stops at about 0.5m from the object.
Then the robot coordinates the whole-body reaching motion and successfully grasps the object.
This visualization indicates that the robot's onboard egocentric view is limited, and the robot can only see the object within a short distance ($<0.6$m), which makes 
the availability of the 3D spatial understanding beforehand~\cite{OKRobot24,VGGT25,ReasoningInSpace26} critical for searching objects, which could be a future exploration.

\begin{figure*}[!h]
  \includegraphics[width=1.0\linewidth]{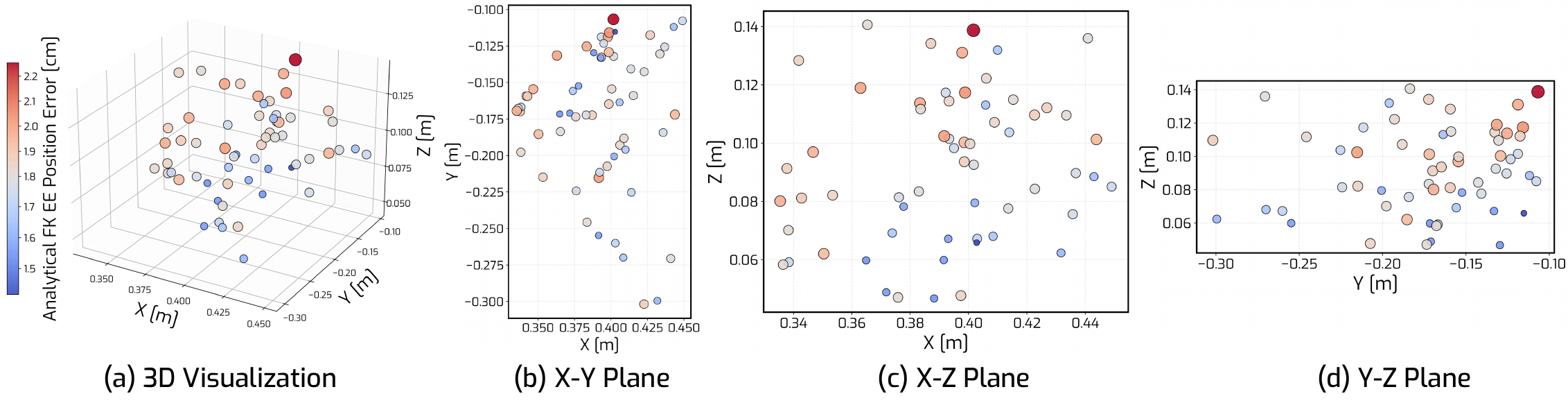}
  \caption{\textbf{Visualization of analytical forward kinematics error}. 
  We plot the 60 data points collected in the MOCAP room.
  The error is indicated in the color bar on the left side of the figure, and the size of the scatter also increases with the error.
  }
  \label{fig:analytical_fk_error_3d_visualization}
\end{figure*}

\begin{figure*}[!t]
  \centering
  \includegraphics[width=1.0\linewidth]{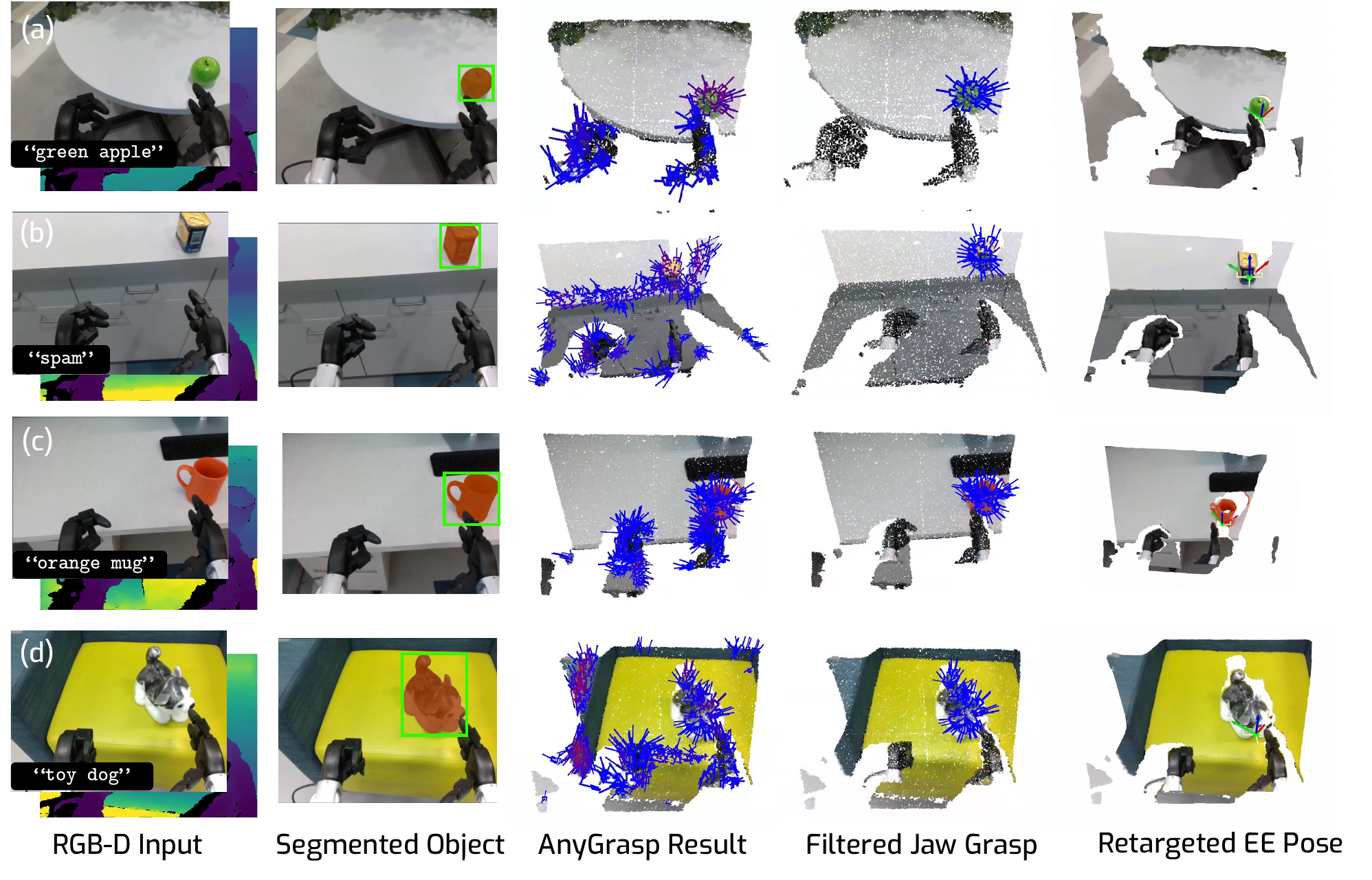}
  \caption{\textbf{Visual perception for language-conditioned grasping.}
  Examples of target segmentation, grasp proposals, filtering, and retargeted EE poses from onboard RGB-D observations.
  }\label{fig:visual_perception2}
\end{figure*}

\subsection{Visual Perception Details}\label{app:perception}
\cref{fig:visual_perception2} illustrates how HERO leverages LVMs to obtain the targeted EE grasping pose, following a modular perception-to-action design that is similar to prior systems~\cite{OKRobot24,SoFAR25}. 

\noindent \textbf{Object of Interest from Language.}
Given an ego-centric RGB-D observation and a natural-language query specifying the target object, HERO first applies GroundingDINO to produce a language-conditioned detection box~\cite{GroundingDINO15_24}. 
The detected box is then used to prompt SAM-3 for segmentation of the object of interest~\cite{SAM325}. 

\noindent\textbf{Grasp Proposals.}
This mask serves as a spatial constraint for grasp proposal generation: HERO runs AnyGrasp~\cite{AnyGrasp23} to produce a set of candidate grasps, and then filters out proposals outside the segmented object region. 
Note that AnyDexGrasp~\cite{AnyDexGrasp25} can also be used here, but we find that the Dex-3 hand lacks dexterity, and the difference between these two methods is limited.

\noindent\textbf{Grasp Selection.}
To select the best jaw grasp, we first filter out the grasp poses that lie on the opposite side of the object relative to the robot's hand (\eg, for an object to the right side of the hand, the left approaching grasps are abandoned).
Then we filter out grasps that are too high or too low based on a gravity-aligned height estimation of objects using depth.
Afterward, we select the grasp that lies most parallel to the ground with the highest confidence as the final grasp.

\noindent\textbf{Grasp Retargeting.}
The selected grasp is retargeted to a 6-DoF end-effector pose for dexterous grasping with the Dex-3 hand. 
We first rotate the gripper pose by 45 degrees around the $z$-axis to improve the grasp robustness and pose error tolerance. After that, we clip the yaw angle within 70 degrees to ensure the orientation is not too large.

\noindent\textbf{Approach-angle sweep.} In a preliminary $0^\circ$--$90^\circ$ sweep, wrist distortion increased from $22^\circ$ to $104^\circ$ as the angle increased, while the observed object knock-over rate fell from $45.1\%$ at $0^\circ$ to $7.4\%$ at $45^\circ$. We use the $45^\circ$ offset as a compromise between these effects.

\subsection{Analytical FK Error Visualization}\label{app:fk-vis}
In \cref{fig:analytical_fk_error_3d_visualization}, we visualize the translation error of analytical forward kinematic results.
We plot the error via the collected 60 samples in the MOCAP room, where the error is recorded when time is 1 minute.
From the figure, we can observe that the error generally increases when the EE location becomes larger along the Y and Z axes, which may form a pattern that can be learned from a neural model.

\subsection{Tracking Error Curves with and without Replanning}
\label{app:execution-curves}
\begin{figure}[!t]
  \includegraphics[width=1.0\linewidth]{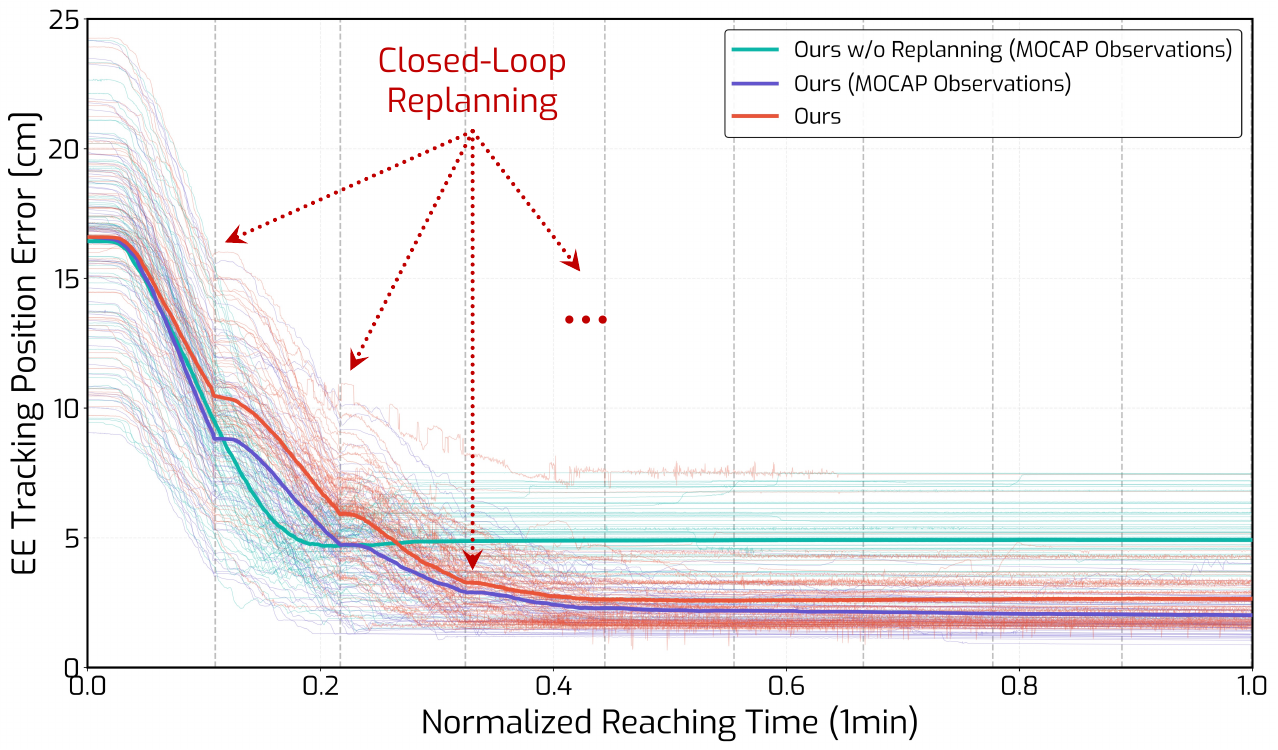}
  \caption{\textbf{Impact of replanning on end-effector tracking error in the real world}. We plot the end-effector translation error as a function of execution time steps. The plot shows 1 minute of execution at 50Hz.
  The transparent lines are individual 60 real-world rollouts, and the corresponding solid line indicates the average value.
  The gray vertical dashlines indicate the replanning every 6 seconds (0.15Hz). \textcolor{curvecolor1}{Cyan} line shows \ours without replanning and while \textcolor{curvecolor2}{purple} line shows \ours with re-planning. Re-planning leads to more accurate tracking. \textcolor{curvecolor3}{Orange} line uses end-effector estimates from our neural model which leads to tracking performance very close to the oracle \textcolor{curvecolor2}{purple} line that uses end-effector estimates from MOCAP.}
  \label{fig:trajectory_analysis}
\end{figure}

\cref{fig:trajectory_analysis} shows how end-effector tracking error evolves over the course of a real-world reach, complementing the aggregate ablations in \cref{tab:tracking-replan}. Without replanning, the error plateaus at a higher level because the policy commits to a stale reference; with replanning, the error continues to decrease as fresh references compensate for accumulated drift. The fact that \ours with learned $\eta,\xi$ closely tracks the MOCAP-oracle curve also confirms that our learned forward models suffice for accurate real-world deployment.

\subsection{Whole-Body Reaching Workspace Analysis}
\label{app:workspace}
We quantify how enabling torso motion via the waist DoFs affects end-effector reachability. We compare two kinematic settings: \textit{i) arms-only}, where IK optimizes the 14-DoF arm joints, and \textit{ii) arms+waist}, where IK additionally optimizes the 3-DoF waist (17 DoFs total).

\noindent\textbf{Workspace estimation.}
We define an axis-aligned 3D candidate region in the robot base frame:
\begin{equation}
x \in [0, 1.0]\ \mathrm{m},\quad
y \in [-1.0, 1.0]\ \mathrm{m},\quad
z \in [-0.5, 1.0]\ \mathrm{m},
\end{equation}
discretized at $0.02\,\mathrm{m}$ resolution. For each grid point $\mathbf{p}$ we attempt IK with cuRobo~\cite{Curobo23} under joint-limit constraints; a point is \emph{reachable} if the solver converges within a fixed iteration budget and the EE position residual lies below a preset tolerance. The workspace volume is approximated by voxel counting,
\begin{equation}
V \approx N_{\mathrm{reach}} \cdot (0.02)^3,
\end{equation}
where $N_{\mathrm{reach}}$ is the number of reachable points.

\noindent\textbf{Effect of waist DoFs.}
\cref{tab:workspace-volume} reports the resulting volumes. Enabling the waist substantially enlarges reachability: the combined two-arm workspace grows from $0.248\,\mathrm{m}^{3}$ to $0.523\,\mathrm{m}^{3}$ ($\sim2.1\times$), and the single-arm workspace grows from $0.166\,\mathrm{m}^{3}$ to $0.426\,\mathrm{m}^{3}$. Bending and twisting the torso effectively repositions the shoulder frame, allowing the EE to cover farther-forward and lower-height targets that are infeasible with a non-actuated waist.
\begin{table}[!t]
\renewcommand{\arraystretch}{0.9}
\setlength{\tabcolsep}{2.0pt}
\caption{\textbf{Reachable workspace volume} across configurations.}
\label{tab:workspace-volume}
\centering
\small
\begin{tabular}{lcc}
\toprule[0.95pt]
\bf Configuration & \bf Single Arm ($\mathrm{m}^3$) & \bf Both Arms ($\mathrm{m}^3$) \\
\midrule[0.6pt]
Arms-only (14 DoFs)      & 0.166 & {0.248} \\
Arms+Waist (17 DoFs)    & \textbf{0.426} & \textbf{0.523} \\
\bottomrule
\end{tabular}
\end{table}

\begin{figure}[!t]
  \includegraphics[width=1.0\linewidth]{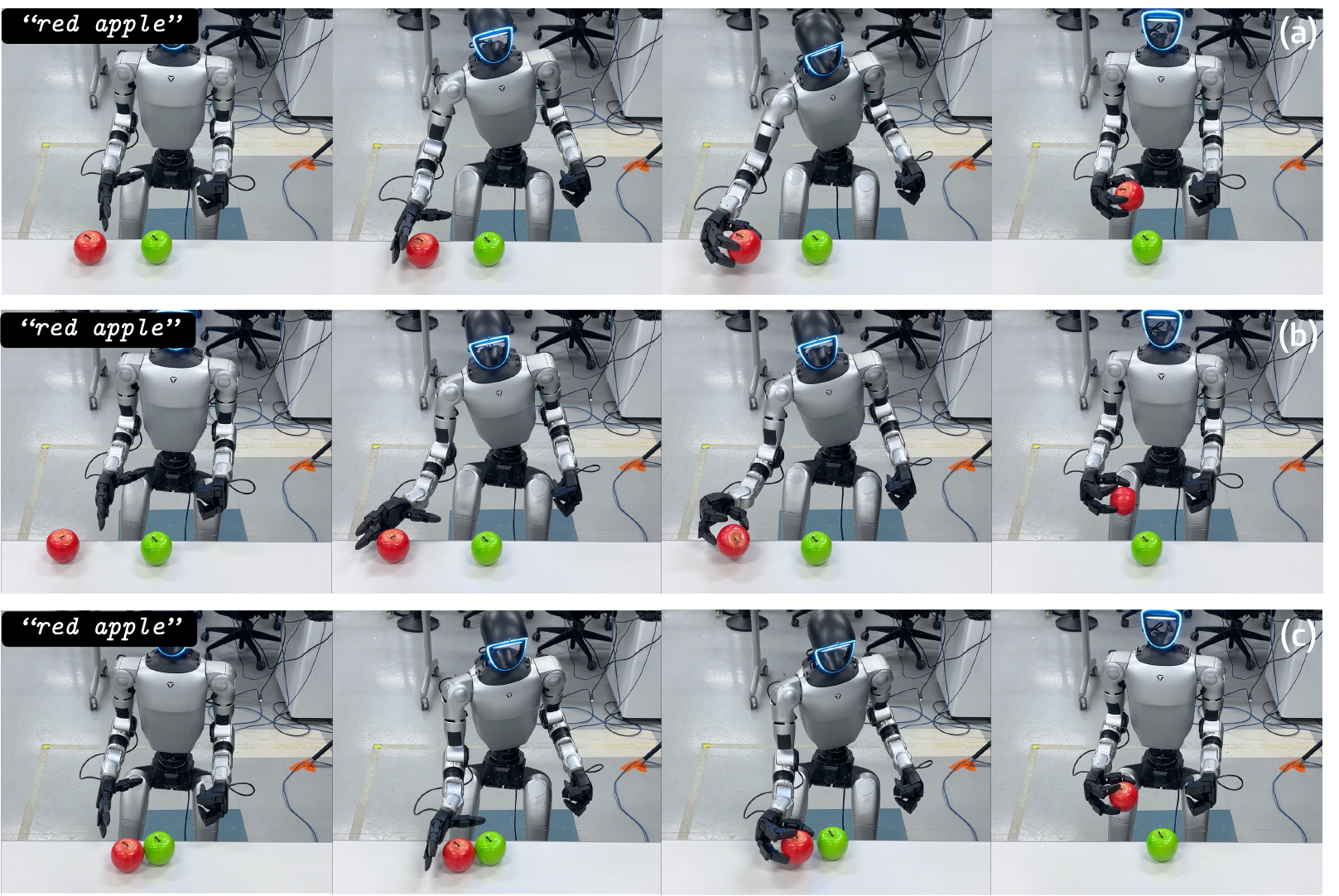}
  \caption{\textbf{HERO enables a humanoid picking up objects from a standard table ($\mathbf{0.74}$m) across a large workspace with open-vocabulary queries.}
    (a-c) The robot can reach and pick up a \texttt{red apple} placed at different heights, poses, and locations.
  }\label{fig:apple_placement1}
\end{figure}

\noindent\textbf{Workspace showcase.}
\cref{fig:apple_placement1} illustrates HERO retrieving objects distributed across an expansive tabletop workspace. Every object is positioned beyond $0.4$\,m from the robot base, so the task requires whole-body coordination to reach precisely while remaining stable. \ours composes expressive whole-body motion with the precision needed for successful grasping.

% ---------------------------------------------------------
% Implementation Details
% ---------------------------------------------------------
\section{Additional Implementation Details}\label{app:impl_details}
\subsection{MOCAP Setup}\label{app:mocap_details}
\noindent\textbf{MOCAP System.} We use the modern MOCAP system Optitrack~\cite{OptitrackSystem} with 13 cameras which provides $\leq0.2\text{mm}$ measure accuracy.

\noindent \textbf{Robot Link Pose.}
To obtain the end-effector pose in the robot frame, we put several markers onto both links, and we show markers on the hand in \cref{fig:mocap_camera2}(a).
Although the MOCAP system provides constructed asset poses via selected marker groups, there exists a misalignment between the MOCAP asset frame and the robot link frame.
To address this, we carefully measure each marker's relative offset to the link's origin, followed by the Kabsch-Umeyama (KU) algorithm~\cite{kabsch1976solution,KULeastSquare02} that transforms individual marker coordinates into 6-DoF link pose in the MOCAP frame within $<1.5\text{mm}$ RMSE error.
The relative transformation of EE and the robot base is thus obtained as they are all in the MOCAP frame.
This approach ensures an accurate measurement of both the end-effector and the robot base, setting a solid ground for our evaluation and camera calibration, introduced next.

\noindent\begin{minipage}{\linewidth}
\textbf{Training-data quantity.} In a separate subsampling study, retraining the learned FK model with 100\%, 50\%, 25\%, and 10\% of a two-hour MOCAP dataset yielded mean EE translation errors of 0.51, 0.61, 0.68, and 0.70~cm, respectively. Base odometry remained usable with 25\% of the data (about 30 minutes).
\end{minipage}

\subsection{Onboard Egocentric RGB-D Camera}\label{app:camera_detail}
\noindent\textbf{Setup.}
We use the onboard RGB-D camera D435i mounted on the humanoid's head, as shown in \cref{fig:mocap_camera2}(b).
The humanoid's neck features a \texttt{pitch} degree of freedom enabling head rotation within a limited range, necessitating precise camera calibration for accurate 3D perception.

\noindent\textbf{Calibration with MOCAP.}
Standard hand-eye calibration~\cite{SolvingHandEyeIterative17,CalibrationOverview12} typically relies on analytical forward kinematics to obtain end-effector poses. 
However, as shown in \cref{tab:forward-kinematics}, analytical forward kinematics has a mean EE translation error of 1.30~cm due to hardware inaccuracies, limiting its suitability for precise camera calibration.

We instead leverage the MOCAP system for ground-truth pose measurement. 
Following the marker-based approach described previously, we attach reflective markers to an ArUco calibration board~\cite{AruCo14} and apply the KU algorithm~\cite{kabsch1976solution,KULeastSquare02} for 6-DoF pose estimation. 
During data collection, we manually move the board through 60-70 diverse poses in front of the camera. 
For each pose $i$, we record: 1) the robot base pose in MOCAP frame $\mathcal{T}_{\text{MOCAP}}^{\text{base}}$, 2) the board pose in MOCAP frame $\mathcal{T}_{\text{MOCAP}}^{\text{board},i}$, and 3) the board pose in camera frame $\mathcal{T}_{\text{camera}}^{\text{board},i}$ via ArUco detection using the OpenCV library~\cite{OpenCV00}.

To compute the camera-to-base transformation $\mathcal{T}_{\text{base}}^{\text{camera}}$, we solve the eye-to-hand calibration problem:
\begin{equation}
\mathcal{T}_{\text{base}}^{\text{camera}} \oplus \mathcal{T}_{\text{camera}}^{\text{board},i} = \mathcal{T}_{\text{MOCAP}}^{\text{base}} \ominus \mathcal{T}_{\text{MOCAP}}^{\text{board},i}
\end{equation}
using the Tsai-Lenz method~\cite{TsaiCalibration85,TsaiLenz89}. 
This MOCAP-assisted calibration achieves a reprojection error within 2.5mm, ensuring accurate egocentric 3D perception.

\noindent \textbf{Image Resolution \& FPS.} We use the RGB-D images with a resolution of $640\times480$ in a 60Hz FPS.

\subsection{Hyper-parameters}
\noindent\textbf{Motion planning}
For motion planning, we use cuRobo and set the planning $dt$ to 7.25e-6.

\noindent\textbf{Grasping Threshold}
When the robot approaches the object, it autonomously close the hand when the hand distance to the target grasp $\DeltaEt\leq \delta$ where $\delta>0$ is a threshold.
At the moment when this threshold is reached, we pass the same local waypoint of the planned motion trajectory to the policy to ensure stability, and the hand is immediately closed for grasping.
In this paper, we utilize a threshold of $\delta=1.5\text{cm}$, which we find most effective across tested objects.

\subsection{Rewards}
\cref{tab:reward} summarizes reward components and weights used for RL training of $\pit$, which is structured into four categories: \textit{tracking task}, \textit{penalties}, \textit{regularization}, and \textit{locomotion task}.
To ensure precise manipulation, the tracking rewards weigh the alignment of the end-effector based on our newly proposed residual $\DeltaEt$.
Note that EE orientation is represented with the continuous 6D parameterization (first two columns of the rotation matrix)~\cite{ContinuityRotation19}. 
To encourage the planned upper-body posture (\eg, waist bending or torso twisting), we also add a joint-space tracking term.
Penalties strictly enforce safety constraints (\eg, joint limits, termination), while regularization terms—such as costs on torque, acceleration, and stance symmetry—are essential for generating smooth, stable motions capable of robust and natural Sim2Real transfer.
To train the robot to follow locomotion commands, we also use a flag variable to control the standing and waking mode switching.
% Draft from the November 2025 reward configuration; verify the evaluated run's Hydra snapshot.
\begingroup
\small
\setlength{\tabcolsep}{3pt}
\renewcommand{\arraystretch}{1.08}
\begin{xltabular}{\linewidth}{@{}p{0.34\linewidth} >{\raggedright\arraybackslash}X r@{}}
\caption{\textbf{Reward terms and nominal weights.}
$m=1$ denotes walking ($m=0$ standing); $h_c$ is commanded base height, $\mathbf g_{xy}$ projected gravity, and $e_p,e_R$ EE position and mean axis-angle errors.
Each weight multiplies the control-step duration; $\dagger$ terms also use a penalty scale $s\in[0,1]$ initialized at $0.1$.
Here $[x]_+=\max(x,0)$, and soft position bounds cover the central 95\% of each joint range.}
\label{tab:reward}\\
\toprule[0.95pt]
{\scshape Term} & {\scshape Expression} & {\scshape Weight} \\
\midrule[0.6pt]
\endfirsthead
\multicolumn{3}{@{}l}{\small\textbf{Table \thetable\ (continued)}}\\
\toprule[0.95pt]
{\scshape Term} & {\scshape Expression} & {\scshape Weight} \\
\midrule[0.6pt]
\endhead
\midrule[0.4pt]
\multicolumn{3}{r}{\footnotesize Continued on next page}\\
\endfoot
\bottomrule[0.95pt]
\endlastfoot

\multicolumn{3}{@{}l}{\textit{\textbf{Tracking task}}}\\
EE position (stance) & $(1-m)\exp(-\|e_p\|_2^2/0.05)$ & $2.0$ \\ % tracking_stance_ee_pos
EE orientation (stance) & $(1-m)\exp(-e_R/0.10)$ & $2.0$ \\ % tracking_stance_ee_rot
Upper-body DoF & $\exp(-\|\mathbf q_{\rm arm}-\mathbf q_{\rm arm}^{\rm ref}\|_2^2/0.1)$ & $4.0$ \\ % tracking_upper_body_dofs
Base height (stance) & $(1-m)\exp(-|h-h_c|/0.05)$ & $4.0$ \\ % tracking_stance_base_height
Base height (walk) & $m\exp(-|h-h_c|/0.05)$ & $1.0$ \\ % tracking_walk_base_height
Waist DoF (stance) & $(1-m)\exp(-\|\mathbf q_{\rm waist}-\mathbf q_{\rm waist}^{c}\|_2^2/0.05)$ & $3.0$ \\ % tracking_waist_dofs_stance
Waist DoF (walk) & $m\exp(-\|\mathbf q_{\rm waist}-\mathbf q_{\rm waist}^{c}\|_2^2/0.05)$ & $0.5$ \\ % tracking_waist_dofs_tapping

\midrule[0.6pt]
\multicolumn{3}{@{}l}{\textit{\textbf{Locomotion task}}}\\
Base velocity $x$ & $\exp(-(v_x-v_x^c)^2/0.25)$ & $2.0$ \\ % tracking_lin_vel_x
Base velocity $y$ & $\exp(-(v_y-v_y^c)^2/0.25)$ & $1.5$ \\ % tracking_lin_vel_y
Base yaw rate & $\exp(-(\omega_z-\omega_z^c)^2/0.25)$ & $4.0$ \\ % tracking_ang_vel
Feet air time & $\begin{aligned}[t]
&\mathds{1}_{\|\mathbf v^c_{xy}\|_2>0.1}\sum_i\mathds{1}_{t_i^{\rm air}>0}\\[-1pt]
&\quad\mathds{1}_{\max(F_{i,z,t},F_{i,z,t-1})>1}(t_i^{\rm air}+\Delta t-0.5)
\end{aligned}$ & $4.0$ \\ % feet_air_time
Foot-heading alignment & $\sum_{i\in\{\rm L,R\}}|\operatorname{wrap}(\psi_i-\psi_{\rm base})|$ & $-0.25$ \\ % feet_heading_alignment

\midrule[0.6pt]
\multicolumn{3}{@{}l}{\textit{\textbf{Regularization}}}\\
Lower-body torque$^\dagger$ & $\|\boldsymbol\tau_{\rm lower}\|_2^2$ & $-10^{-5}$ \\ % penalty_lower_body_torques
Upper-body torque$^\dagger$ & $\|\boldsymbol\tau_{\rm upper}\|_2^2$ & $-10^{-5}$ \\ % penalty_upper_body_torques
Lower-body DoF velocity$^\dagger$ & $\|\dot{\mathbf q}_{\rm lower}\|_2^2$ & $-10^{-3}$ \\ % penalty_lower_body_dof_vel
Upper-body DoF velocity$^\dagger$ & $\|\dot{\mathbf q}_{\rm upper}\|_2^2$ & $-10^{-3}$ \\ % penalty_upper_body_dof_vel
Lower-body DoF acceleration$^\dagger$ & $\|(\dot{\mathbf q}_{t}-\dot{\mathbf q}_{t-1})_{\rm lower}/\Delta t\|_2^2$ & $-2.5\!\times\!10^{-7}$ \\ % penalty_lower_body_dof_acc
Upper-body DoF acceleration$^\dagger$ & $\|(\dot{\mathbf q}_{t}-\dot{\mathbf q}_{t-1})_{\rm upper}/\Delta t\|_2^2$ & $-2.5\!\times\!10^{-7}$ \\ % penalty_upper_body_dof_acc
Lower-body action rate$^\dagger$ & $\|\mathbf a_{{\rm lower},t}-\mathbf a_{{\rm lower},t-1}\|_2^2$ & $-0.1$ \\ % penalty_lower_body_action_rate
Upper-body action rate$^\dagger$ & $\|\mathbf a_{{\rm upper},t}-\mathbf a_{{\rm upper},t-1}\|_2^2$ & $-0.1$ \\ % penalty_upper_body_action_rate
EE linear-velocity change & $\sum_{i\in\{\rm L,R\}}\|\mathbf v_{i,t}^{\rm EE}-\mathbf v_{{\rm L},t-1}^{\rm EE}\|_2$ & $-0.2$ \\ % penalty_ee_lin_acc
EE angular-velocity change & $\sum_{i\in\{\rm L,R\}}\|\boldsymbol\omega_{i,t}^{\rm EE}-\boldsymbol\omega_{{\rm R},t-1}^{\rm EE}\|_2$ & $-0.02$ \\ % penalty_ee_ang_acc
% The historical code cached both linear histories from the left hand and both angular histories from the right hand.
Base vertical velocity & $m(v_z^{\rm base})^2$ & $-2.0$ \\ % penalty_lin_vel_z
Base angular velocity & $\|\boldsymbol\omega^{\rm base}_{xy}\|_2^2$ & $-0.05$ \\ % penalty_ang_vel_xy
Torso angular velocity & $\|\boldsymbol\omega^{\rm torso}_{xy}\|_2^2$ & $-1.0$ \\ % penalty_ang_vel_xy_torso
Base orientation & $\|\mathbf g^{\rm base}_{xy}\|_2^2$ & $-1.5$ \\ % penalty_orientation
Torso orientation & $\|\mathbf g^{\rm torso}_{xy}\|_1$ & $-1.0$ \\ % penalty_torso_orientation
Base-height deviation & $(5-4m)(h-h_c)^2$ & $-10.0$ \\ % base_height
Hip roll/yaw position & $[m+(1-m)h_c]\,\|\mathbf q_{\rm hip,roll/yaw}\|_2^2$ & $-2.5$ \\ % penalty_hip_pos
Stance DoF velocity & $(1-m)\|\dot{\mathbf q}_{\rm lower}\|_2^2$ & $-10^{-3}$ \\ % penalty_stance_dof
Stance symmetry & $(1-m)(\|\mathbf q_{\rm L}^{s}-\mathbf q_{\rm R}^{s}\|_1+\|\mathbf q_{\rm L}^{a}+\mathbf q_{\rm R}^{a}\|_1)$ & $-0.5$ \\ % penalty_stance_symmetry
Ankle roll & $|q_{\rm ankle,L}|+|q_{\rm ankle,R}|$ & $-2.0$ \\ % penalty_ankle_roll
Stance root offset & $\begin{aligned}[t]
&(1-m)|[R_{\rm pel}^{\top}(\mathbf p_{\rm pel}\\[-1pt]
&\quad-\tfrac12(\mathbf p_{\rm L}+\mathbf p_{\rm R}))]_y|
\end{aligned}$ & $-5.0$ \\ % penalty_stance_root
Stance foot offset & $\begin{aligned}[t]
&(1-m\mathds{1}_{|v_x^c|>0})\\[-1pt]
&\quad|[R_{\rm pel}^{\top}\operatorname{abs}(\mathbf p_{\rm L}-\mathbf p_{\rm R})]_x|
\end{aligned}$ & $-5.0$ \\ % penalty_stance_tap_feet
Zero-command translation & $m\,\mathds{1}_{\|\mathbf v_{xy}^{c}\|<0.2}\|\mathbf v_{xy}^{\rm root,world}\|_2$ & $-1.5$ \\ % penalty_shift_in_zero_command
Zero-command rotation & $m\,\mathds{1}_{|\omega_z^{c}|<0.1}|\omega_z^{\rm root,world}|$ & $-1.5$ \\ % penalty_ang_shift_in_zero_command
Contact slip & $\begin{aligned}[t]
&(10-9m)\sum_i\mathds{1}_{\|\mathbf F_i\|_2>1}\\[-1pt]
&\quad\|\mathbf v_i^{\rm foot}\|_2^2
\end{aligned}$ & $-0.2$ \\ % penalty_contact_no_vel
Foot-contact pattern & $\begin{aligned}[t]
&(1-m)\mathds{1}_{n_f<2}+m\mathds{1}_{n_f\ne1},\\[-1pt]
&n_f=\sum_i\mathds{1}_{F_{i,z}>1}
\end{aligned}$ & $-4.0$ \\ % penalty_contact
Standing foot contact & $(1-m)\mathds{1}_{\min_i F_{i,z}<0.1}$ & $-0.15$ \\ % penalty_stand_still
Feet orientation & $\sum_{i\in\{\rm L,R\}}\|\mathbf g^i_{xy}\|_2$ & $-2.0$ \\ % penalty_feet_ori
Foot target height & $m[\min_i|z_i-0.11|-0.02]_+$ & $-5.0$ \\ % penalty_feet_height
Swing-foot height & $\begin{aligned}[t]
&\sum_{i:\|\mathbf F_i\|_2\le1}\\[-1pt]
&\quad(z_i-0.11m-0.025(1-m))^2
\end{aligned}$ & $-20.0$ \\ % penalty_feet_swing_height
Feet too close & $\mathds{1}_{\|\mathbf p_{\rm L}^{xy}-\mathbf p_{\rm R}^{xy}\|_2<0.17}$ & $-10.0$ \\ % penalty_close_feet_xy
Knee below threshold & $\sum_{i\in\{\rm L,R\}}\mathds{1}_{q_{{\rm knee},i}<0.2}$ & $-1.0$ \\ % penalty_negative_knee_joint
Feet air-time asymmetry & $|t_{\rm L}^{\rm last}-t_{\rm R}^{\rm last}|$ & $-5.0$ \\ % penalty_diff_feet_air_time

\midrule[0.6pt]
\multicolumn{3}{@{}l}{\textit{\textbf{Penalties}}}\\
Lower-body DoF position limit$^\dagger$ & $\sum_{j\in{\rm lower}}\operatorname{dist}(q_j,[q_{j,\min}^{\rm soft},q_{j,\max}^{\rm soft}])$ & $-5.0$ \\ % limits_lower_body_dof_pos
Upper-body DoF position limit$^\dagger$ & $\sum_{j\in{\rm upper}}\operatorname{dist}(q_j,[q_{j,\min}^{\rm soft},q_{j,\max}^{\rm soft}])$ & $-5.0$ \\ % limits_upper_body_dof_pos
Upper-body DoF velocity limit$^\dagger$ & $\sum_{j\in{\rm upper}}\min\{1,[|\dot q_j|-\dot q_{j,\max}]_+\}$ & $-5.0$ \\ % limits_upper_body_dof_vel
Upper-body torque limit$^\dagger$ & $\sum_{j\in{\rm upper}}[|\tau_j|-0.9\tau_{\max,j}]_+$ & $-0.1$ \\ % limits_upper_body_torque
Termination & $b_{\rm reset}(1-b_{\rm timeout})$ & $-250$ \\ % termination
\end{xltabular}
\endgroup

\subsection{Policy Training}
\noindent\textbf{Simulation \& Training Setup.}
We train our end-effector tracking policy $\pit$ in Isaac Gym~\cite{IsaacGym21} or Isaac Sim via Isaac Lab~\cite{mittal2025isaac}, and transfer this policy to the MuJoco simulator~\cite{Mujoco12} for Sim2Sim evaluations before deploying it in the real world.
We train our policy with 4,096 environments for overall $20$K iterations in parallel, with a learning rate of 1e-4 for both the actor and critic models.
AdamW optimizer~\cite{AdamW19} is used with a weight decay of 1e-2.
We use a high simulation frequency of 500Hz, with the low-level PD controller running at 50Hz.
All the policy training is conducted on a single NVIDIA RTX 4090 or an L40S GPU.

\noindent\textbf{Sim2Real Domain Randomization.}
Following previous works~\cite{OmniH2O24,ASAP25,HumanUP25}, we employ standard domain and dynamics randomization to facilitate Sim2Real transfer~\cite{DomainRandomizationSim2Real17}, including variations in link center of mass (CoM) and control delay. 
Notably, we identify that randomizing the \textit{end-effector mass} is essential; without this specific randomization, the policy exhibits end-effector instability, leading to high-frequency hand oscillations that compromise tracking accuracy.

\subsection{Deployment Hardware}
We run all modules (\eg, $\pit$ and SAM-3~\cite{SAM325}) off-the-shelf on a 32-GB RAM laptop equipped with NVIDIA RTX 5070Ti GPU and Intel Core Ultra 9 275HX CPU processor (24 CPU cores / 24 threads).
We run cuRobo with CUDA graph acceleration, which largely improve the efficiency on the edge~\cite{Curobo23}.
For the detection module, we have tested both Grounding DINO base~\cite{GroundingDINO24} and Grounding DINO 1.5~\cite{GroundingDINO15_24},
where the base version can be deployed on the laptop, and Grounding DINO 1.5 only provides access through online APIs.
However, we find that Grounding DINO base is sufficient for most scenes and objects.
\cref{tab:runtime-breakdown} reports deployment runtime; conservative robot motion accounts for most of the roughly 55-second execution.
\begin{table}[H]
\caption{\textbf{Deployment runtime.} Measured on the local laptop with Grounding DINO base. GroundingSAM and AnyGrasp are included in perception; replanning is a per-update latency and is not an additional phase in the total.}
\label{tab:runtime-breakdown}
\centering
\small
\begin{tabular}{lr}
\toprule
Phase or component & Time \\
\midrule
Perception (query to grasp) & 1.33 s \\
\quad GroundingSAM & 950 ms \\
\quad AnyGrasp & 377 ms \\
Reaching & 20.39 s \\
Closure & 1.20 s \\
Return & 32.14 s \\
\midrule
Total (four phases) & $\approx 55$ s \\
\midrule
Replanning (per update) & $\approx 20$ ms \\
\bottomrule
\end{tabular}
\end{table}

\subsection{Testing Assets Details}\label{app:testing_assets}
In the paper, we have tested HERO with 20 daily objects; these objects have different sizes and weights, while being made with different materials, making it challenging to grasp with a Dex-3 hand.
We list the detailed sizes, weights, materials, and language queries of all objects tested in \cref{tab:testing-objects-details-merged}.
Note that the size is roughly measured as the shape is irregular and cannot be easily described.

\subsection{Testing Scenes Details}
\label{app:testing_scenes}
\cref{tab:testing-scenes-details} lists the details of the novel scenes tested in this paper, which are mainly chosen from Coordinated Science
Laboratory Studio (CSL Studio) and the Thomas M. Siebel
Center for Computer Science at the University of Illinois
Urbana-Champaign, Urbana, IL.
The snapshot of these testing scenes can be found in \cref{fig:benchmark_generalization}.
\clearpage
\onecolumn

\setlength{\tabcolsep}{4pt}
\renewcommand{\arraystretch}{1.15}

\begin{xltabular}{\textwidth}{@{}c p{3.1cm} p{1.6cm} >{\raggedright\arraybackslash}X >{\raggedright\arraybackslash}X@{}}
\caption{\textbf{Testing objects, sizes, weights, materials, and language queries.}
Sizes are roughly measured due to irregular shapes. Weights are measured with an accurate food scale.}
\label{tab:testing-objects-details-merged}\\

\toprule
\textbf{Object} & \textbf{Size} & \textbf{Weight} & \textbf{Material} & \textbf{Language Query} \\
\midrule
\endfirsthead

\toprule
\textbf{Object} & \textbf{Size} & \textbf{Weight} & \textbf{Material} & \textbf{Language Query} \\
\midrule
\endhead

\midrule
\multicolumn{5}{r}{\small Continued on next page} \\
\endfoot

\bottomrule
\endlastfoot

\multicolumn{5}{@{}l}{\textit{10 daily object evaluation.}}\\
\midrule

\begin{tabular}{@{}c@{}}
\includegraphics[width=0.05\textwidth]{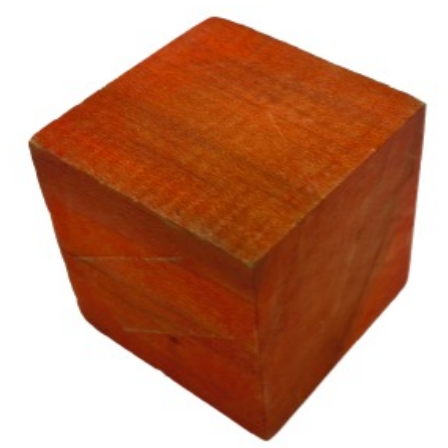}
\end{tabular}
& \small 4.9$\times$4.9$\times$4.9\,cm
& \small 58.06\,g
& \small Wood
& \small \texttt{orange cube}
\\

\begin{tabular}{@{}c@{}}
\includegraphics[width=0.05\textwidth]{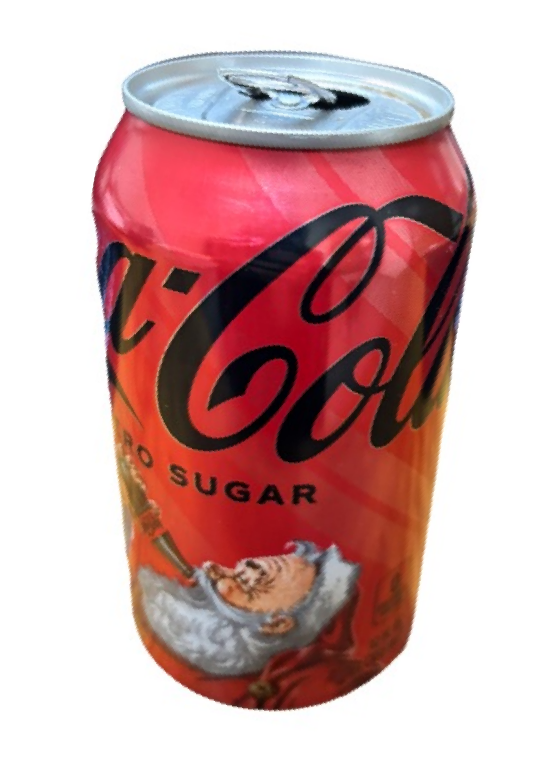}
\end{tabular}
& \small 12$\times$6$\times$6\,cm
& \small 14.97\,g
& \small Aluminum
& \small \texttt{coke can}
\\

\begin{tabular}{@{}c@{}}
\includegraphics[width=0.05\textwidth]{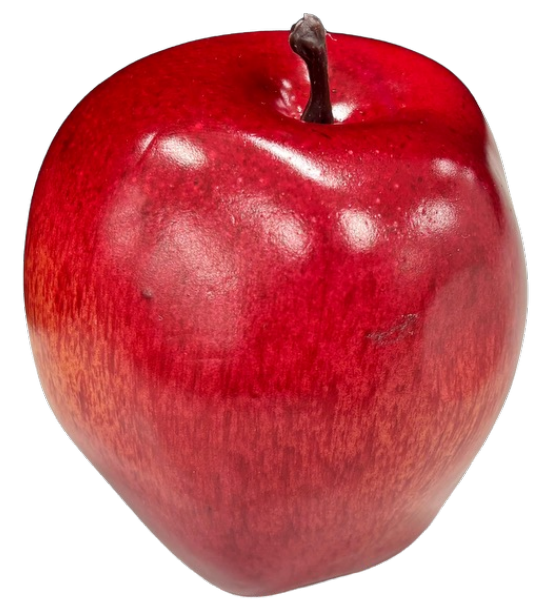}
\end{tabular}
& \small 8.5$\times$7.5$\times$7.5\,cm
& \small 15.88\,g
& \small Plastic
& \small \texttt{red apple}
\\

\begin{tabular}{@{}c@{}}
\includegraphics[width=0.05\textwidth]{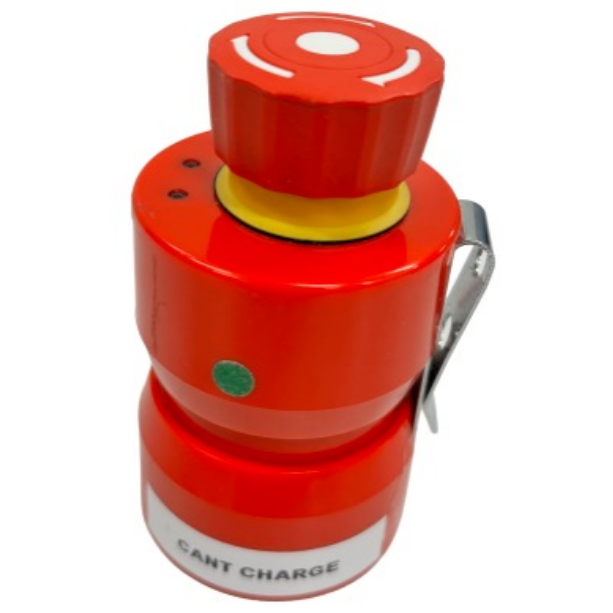}
\end{tabular}
& \small 9$\times$5$\times$5\,cm
& \small 137.89\,g
& \small Plastic \& Metal
& \small \texttt{emergency stop button}
\\

\begin{tabular}{@{}c@{}}
\includegraphics[width=0.05\textwidth]{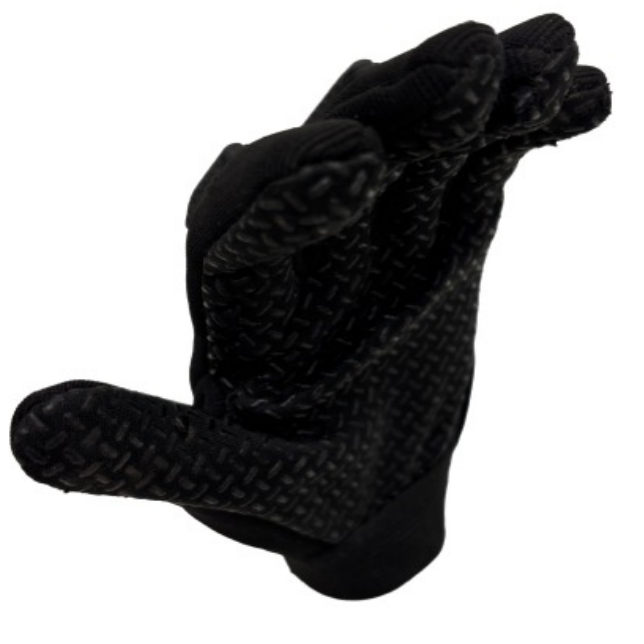}
\end{tabular}
& \small 15$\times$6$\times$5\,cm
& \small 239.95\,g
& \small Plastic \& Metal
& \small \texttt{robot hand}
\\

\begin{tabular}{@{}c@{}}
\includegraphics[width=0.05\textwidth]{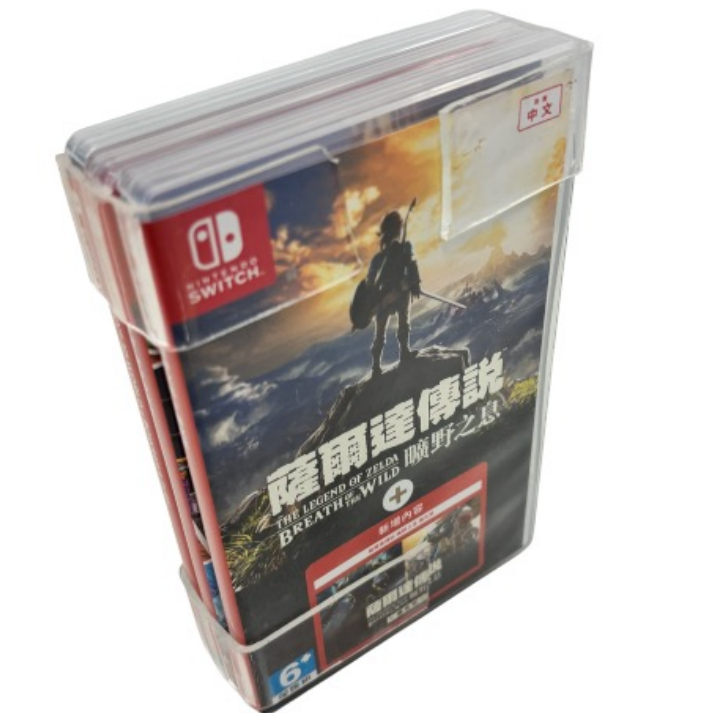}
\end{tabular}
& \small 16.5$\times$10.5$\times$3.3\,cm
& \small 185.07\,g
& \small Plastic 
& \small \texttt{game cartridge}
\\

\begin{tabular}{@{}c@{}}
\includegraphics[width=0.05\textwidth]{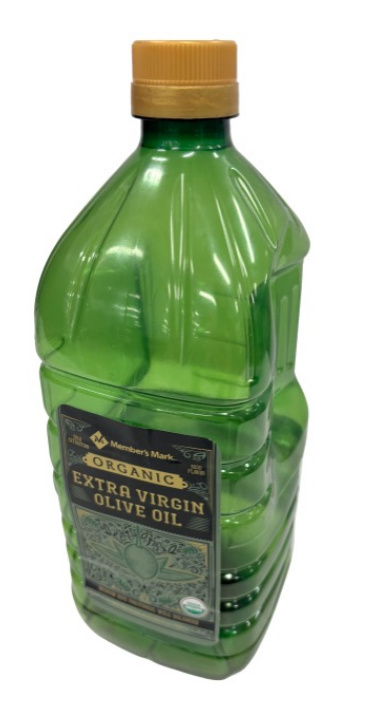}
\end{tabular}
& \small 27.5$\times$9.5$\times$9.5\,cm
& \small 79.83\,g
& \small Plastic 
& \small \texttt{olive oil bottle}
\\

\begin{tabular}{@{}c@{}}
\includegraphics[width=0.05\textwidth]{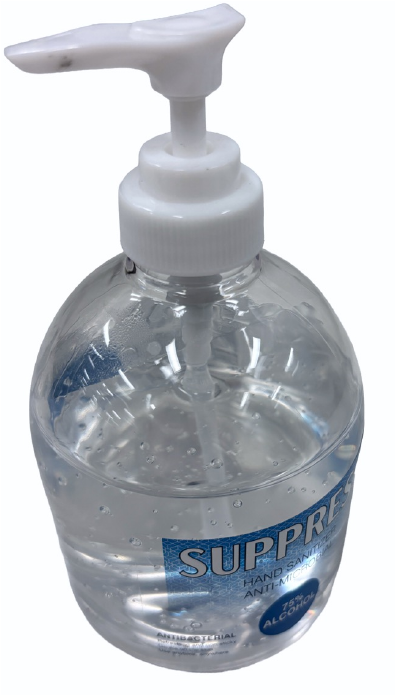}
\end{tabular}
& \small 16$\times$8.5$\times$8.5\,cm
& \small 392.81\,g
& \small Plastic \& Liquid
& \small \texttt{hand soap}
\\

\begin{tabular}{@{}c@{}}
\includegraphics[width=0.05\textwidth]{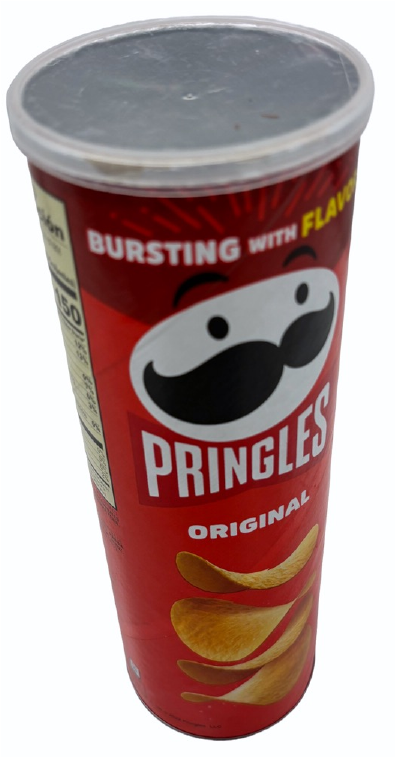}
\end{tabular}
& \small 23$\times$7.9$\times$7.9\,cm
& \small 43.09\,g
& \small Paperboard \& Plastic
& \small \texttt{chip can}
\\

\begin{tabular}{@{}c@{}}
\includegraphics[width=0.05\textwidth]{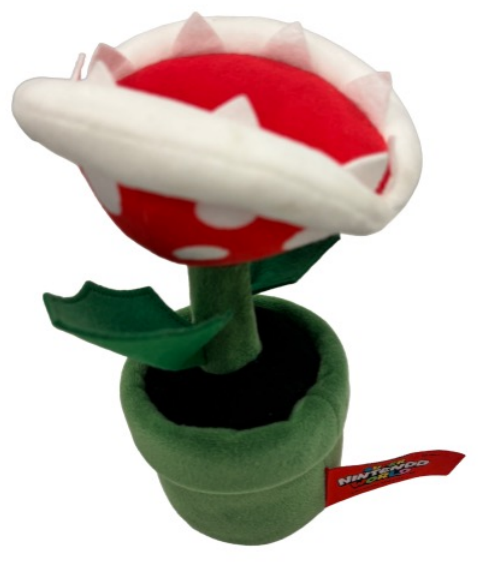}
\end{tabular}
& \small 17$\times$11$\times$11\,cm
& \small 73.94\,g
& \small Plush
& \small \texttt{red piranha plant}
\\

\addlinespace[2pt]
\midrule
\multicolumn{5}{@{}l}{\textit{10 daily objects used in 10 daily scenes evaluation.}}\\
\midrule

% -------------------- Group 2 --------------------
\begin{tabular}{@{}c@{}}
\includegraphics[width=0.05\textwidth]{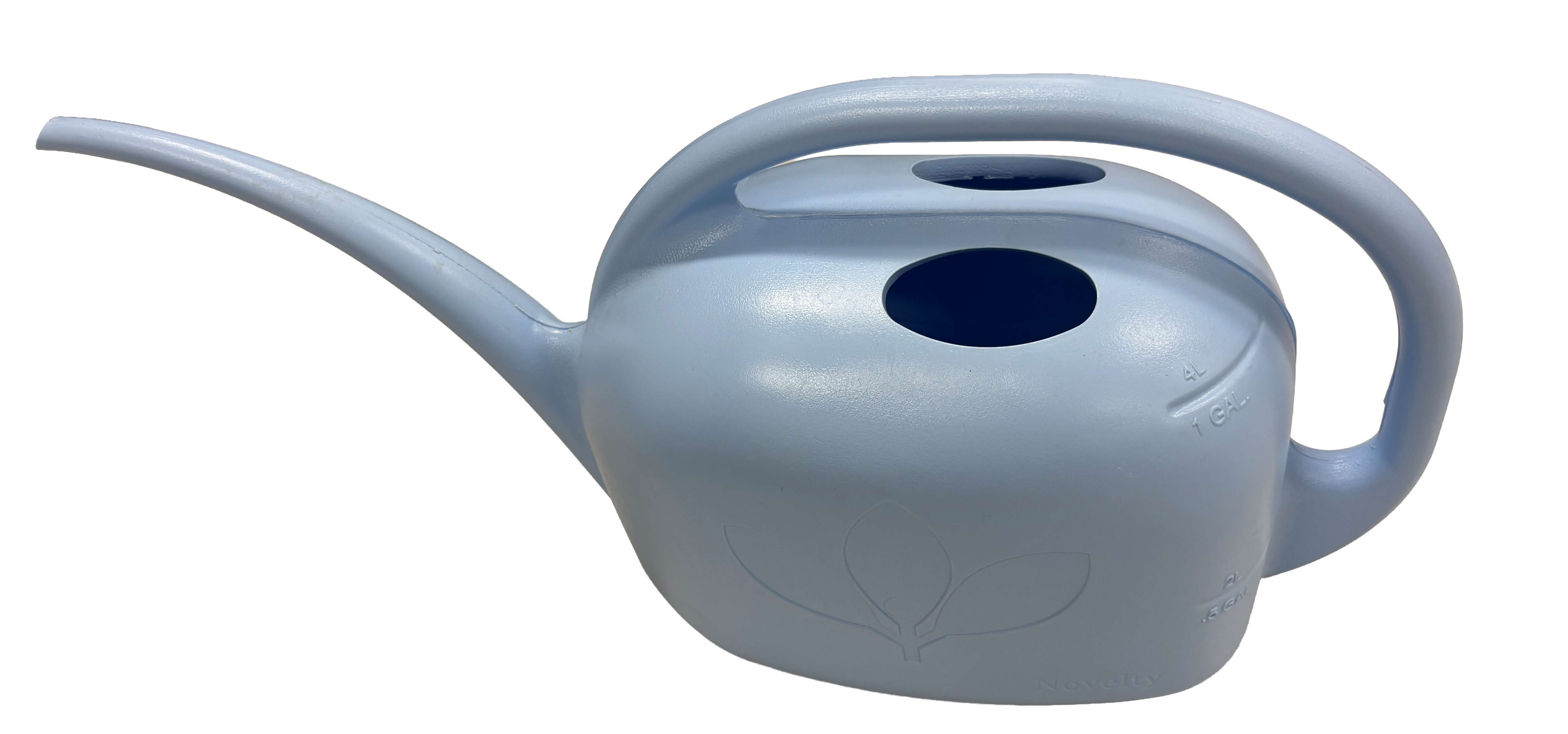}
\end{tabular}
& \small 21$\times$12$\times$43\,cm
& \small 215.91\,g
& \small Plastic
& \small \texttt{kettle}
\\

\begin{tabular}{@{}c@{}}
\includegraphics[width=0.05\textwidth]{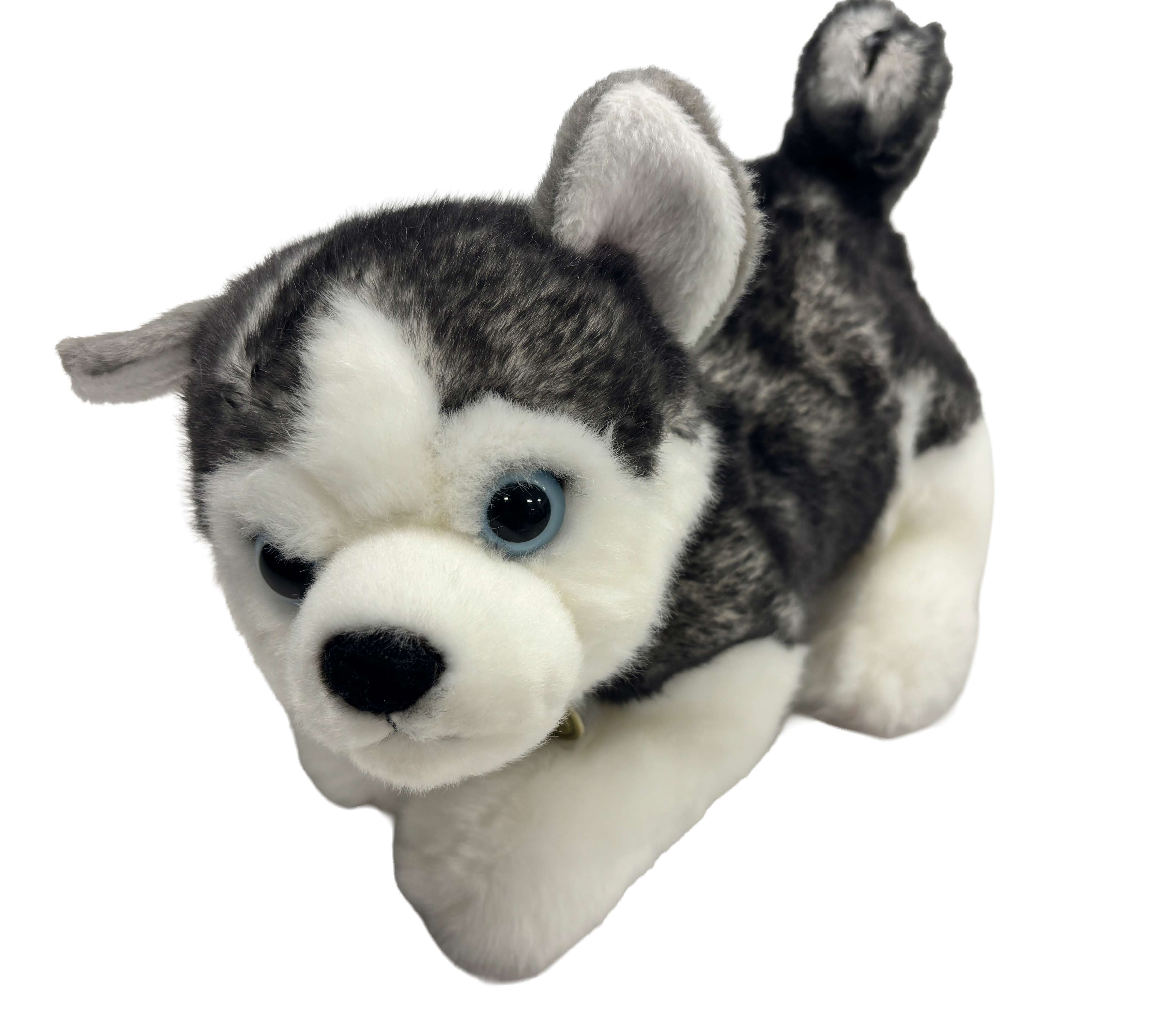}
\end{tabular}
& \small 16.5$\times$11$\times$21\,cm
& \small 213.19\,g
& \small Plush
& \small \texttt{toy dog}
\\

\begin{tabular}{@{}c@{}}
\includegraphics[width=0.05\textwidth]{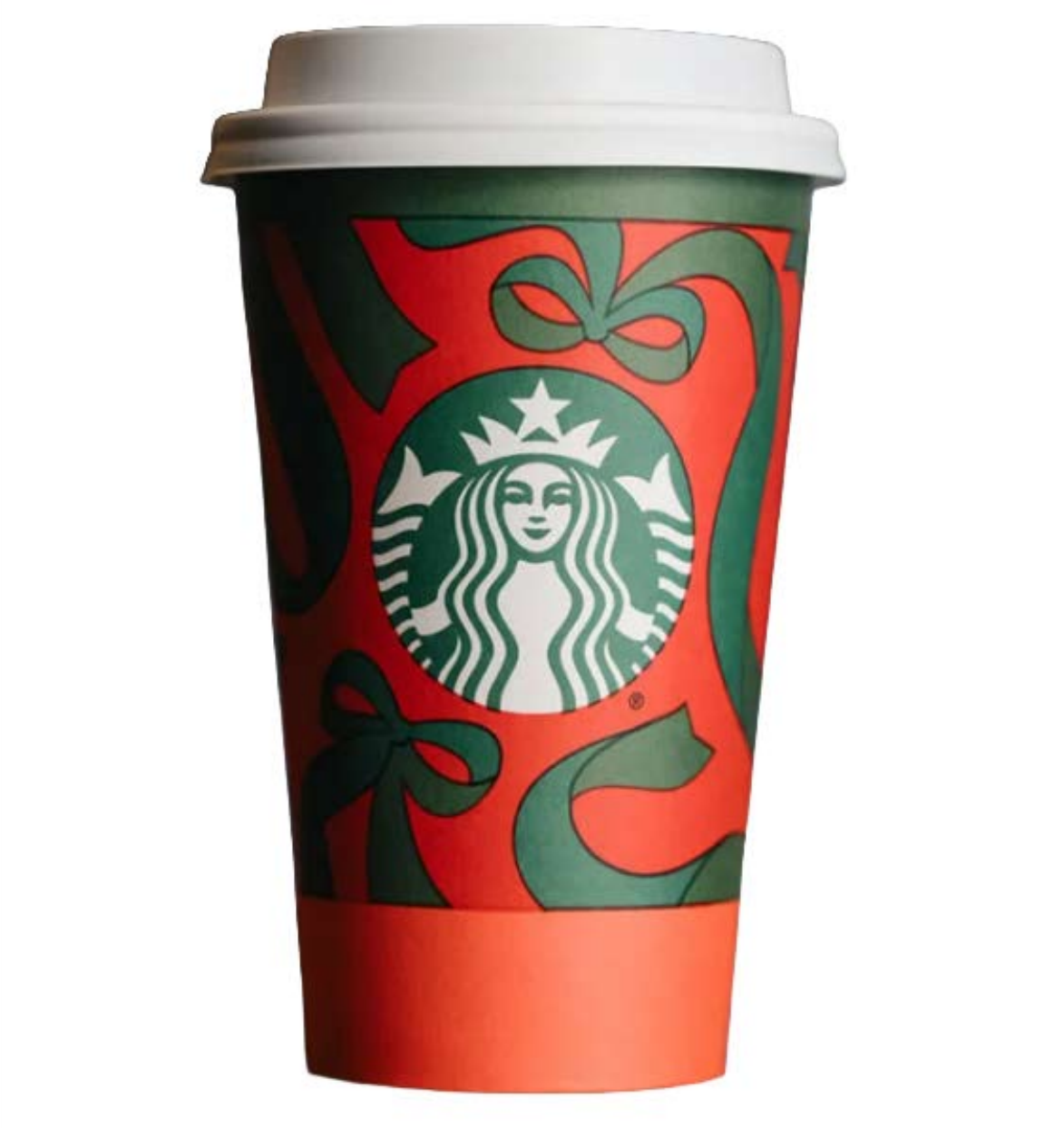}
\end{tabular}
& \small 17.2$\times$9$\times$9\,cm
& \small 24.95\,g
& \small Paperboard \& Plastic
& \small \texttt{Starbucks coffee}
\\

\begin{tabular}{@{}c@{}}
\includegraphics[width=0.05\textwidth]{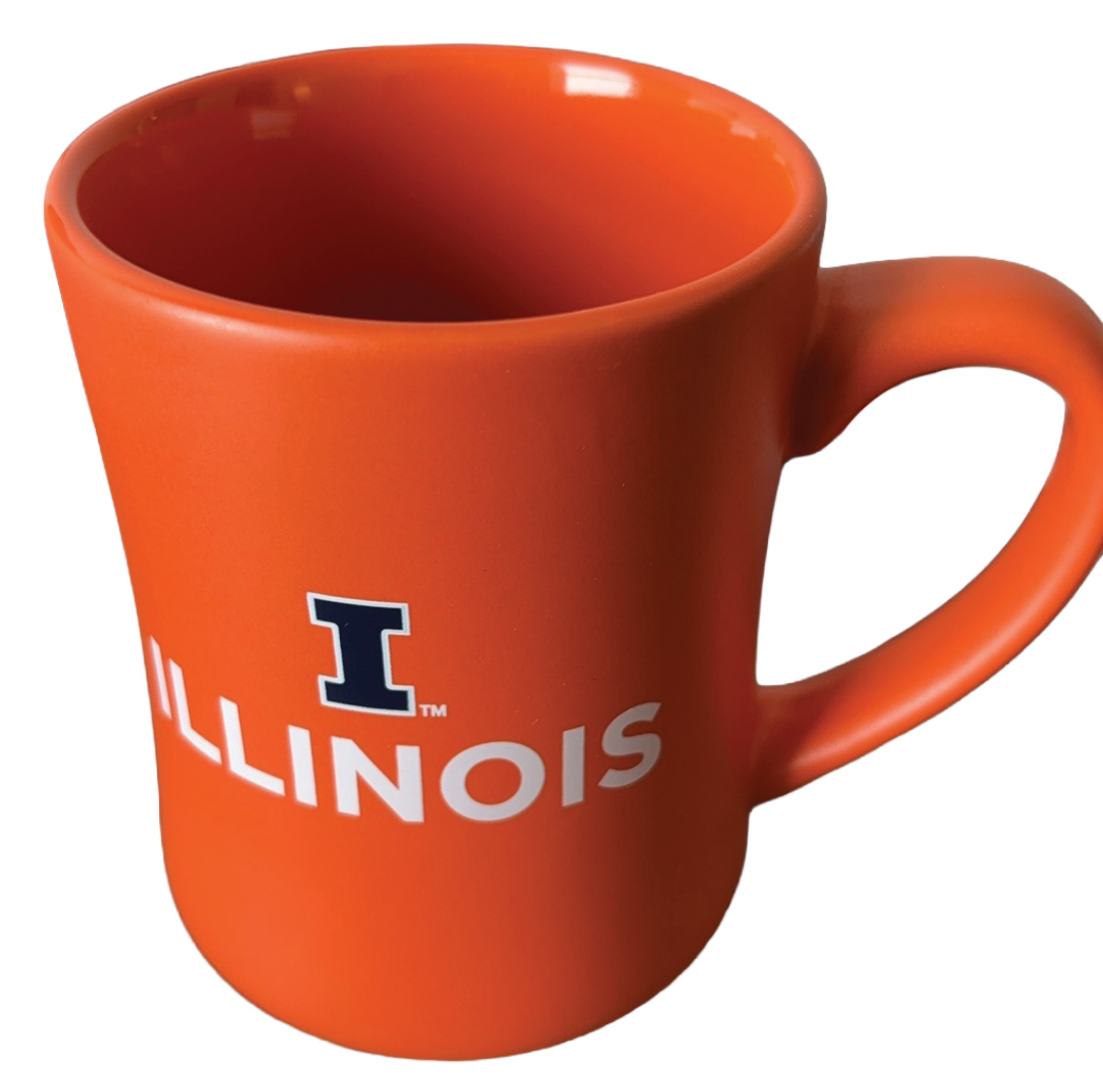}
\end{tabular}
& \small 11.5$\times$9.4$\times$9.4\,cm
& \small 526.17\,g
& \small Ceramic
& \small \texttt{orange mug}
\\

\begin{tabular}{@{}c@{}}
\includegraphics[width=0.05\textwidth]{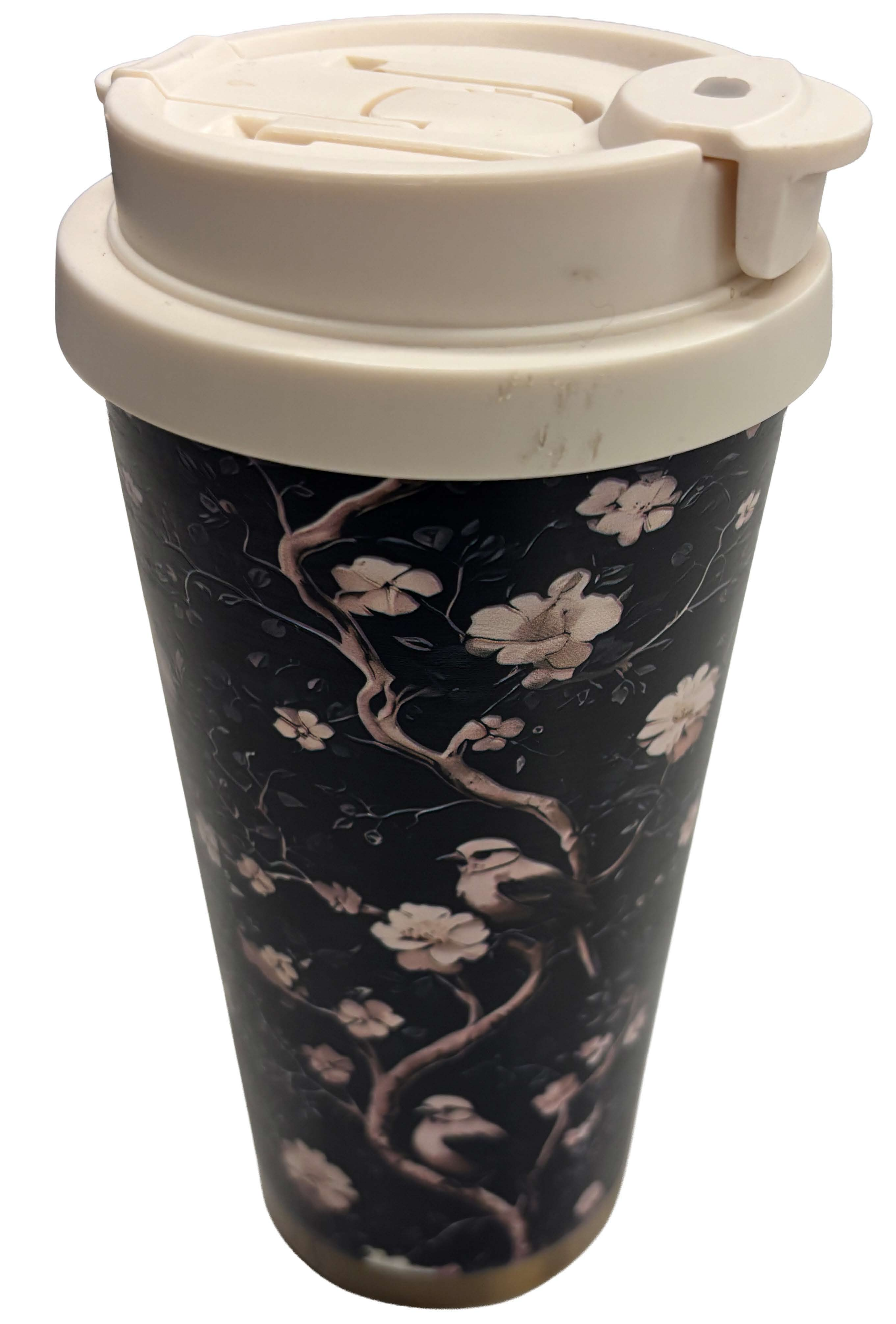}
\end{tabular}
& \small 18.5$\times$6.3$\times$6.3\,cm
& \small 286.22\,g
& \small Plastic \& Metal
& \small \texttt{water bottle}
\\

\begin{tabular}{@{}c@{}}
\includegraphics[width=0.05\textwidth]{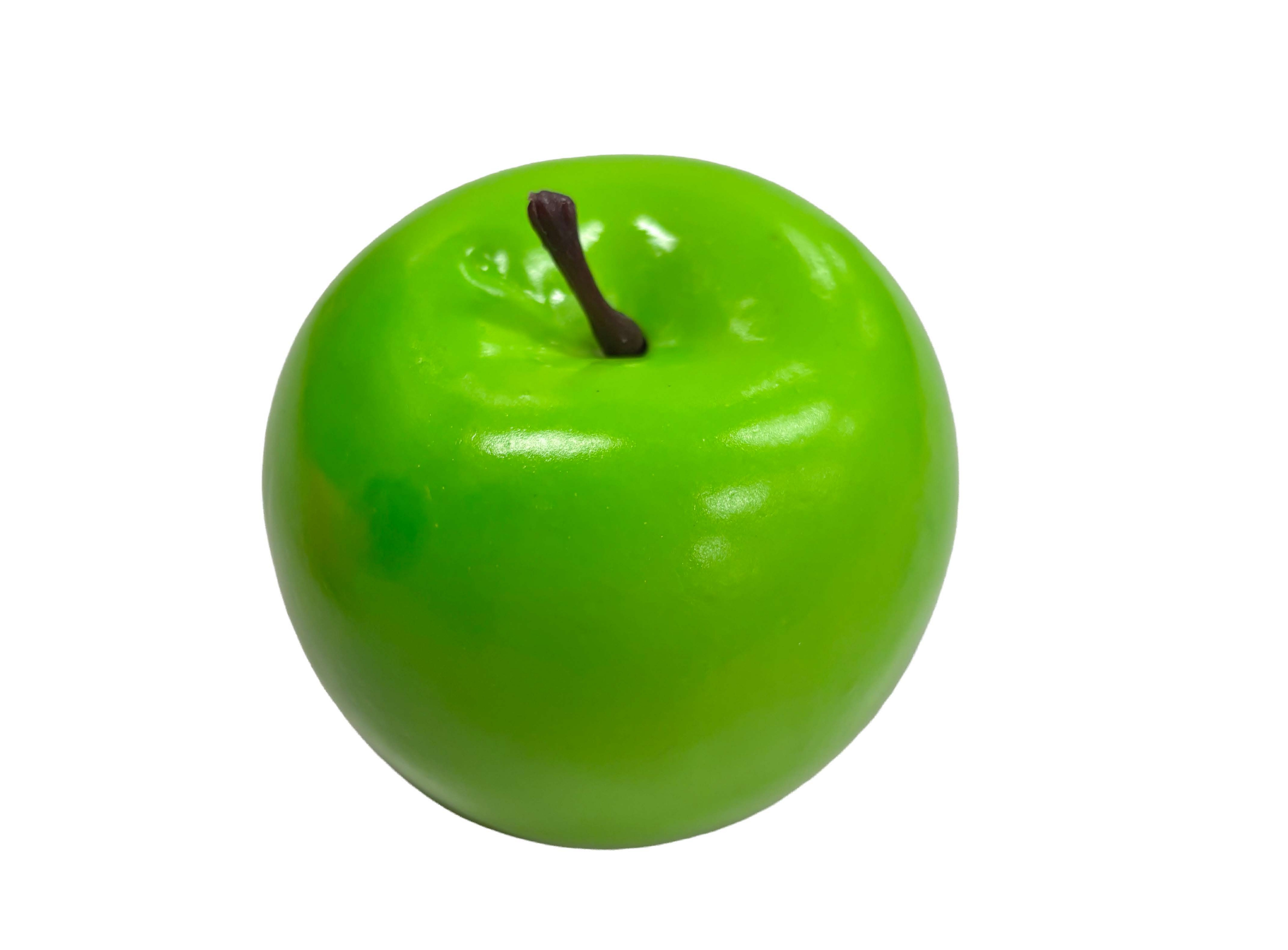}
\end{tabular}
& \small 7$\times$7.7$\times$7.7\,cm
& \small 14.97\,g
& \small Plastic 
& \small \texttt{green apple}
\\

\begin{tabular}{@{}c@{}}
\includegraphics[width=0.05\textwidth]{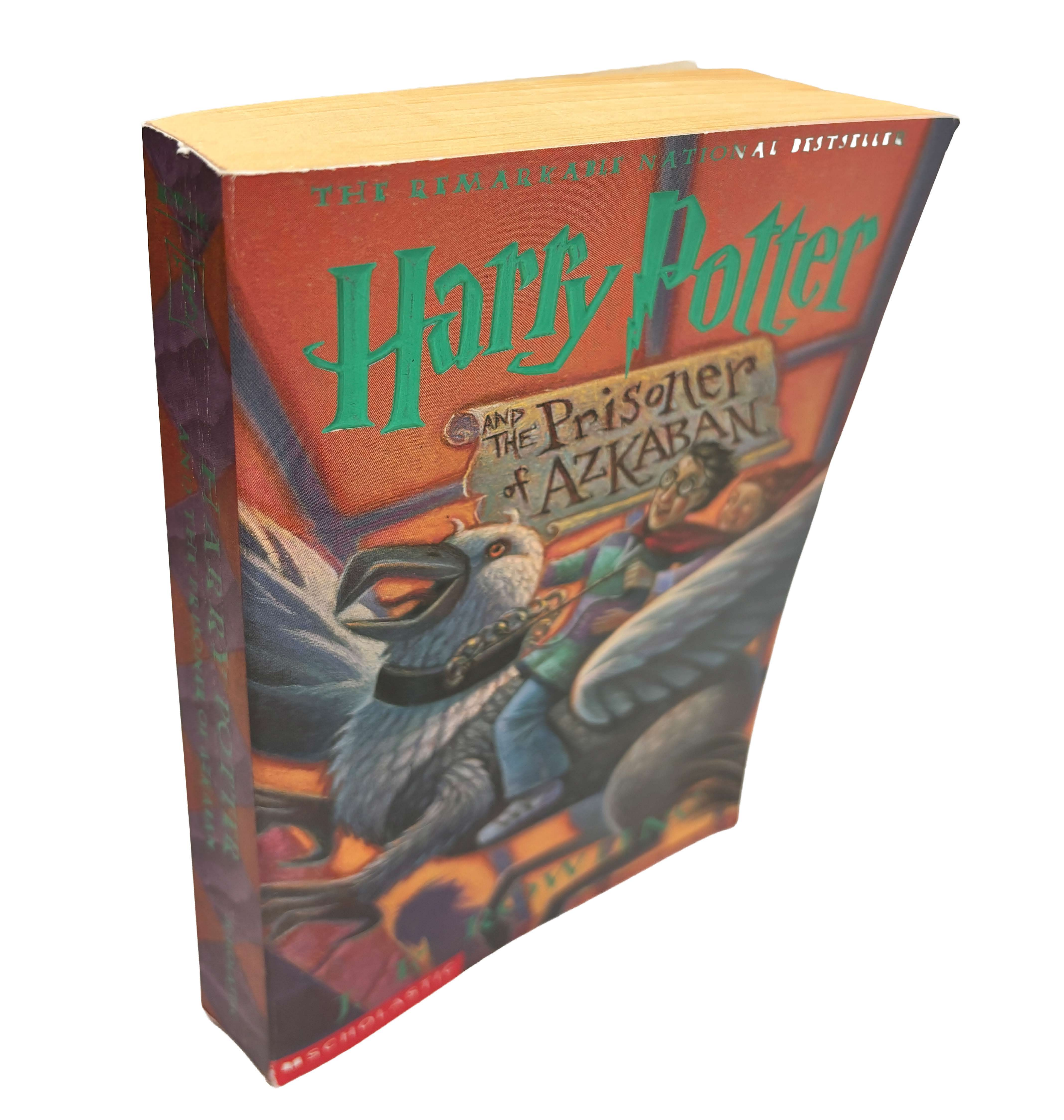}
\end{tabular}
& \small 18.8$\times$2.4$\times$13.3\,cm
& \small 301.19\,g
& \small Paper
& \small \texttt{purple book}
\\

\begin{tabular}{@{}c@{}}
\includegraphics[width=0.05\textwidth]{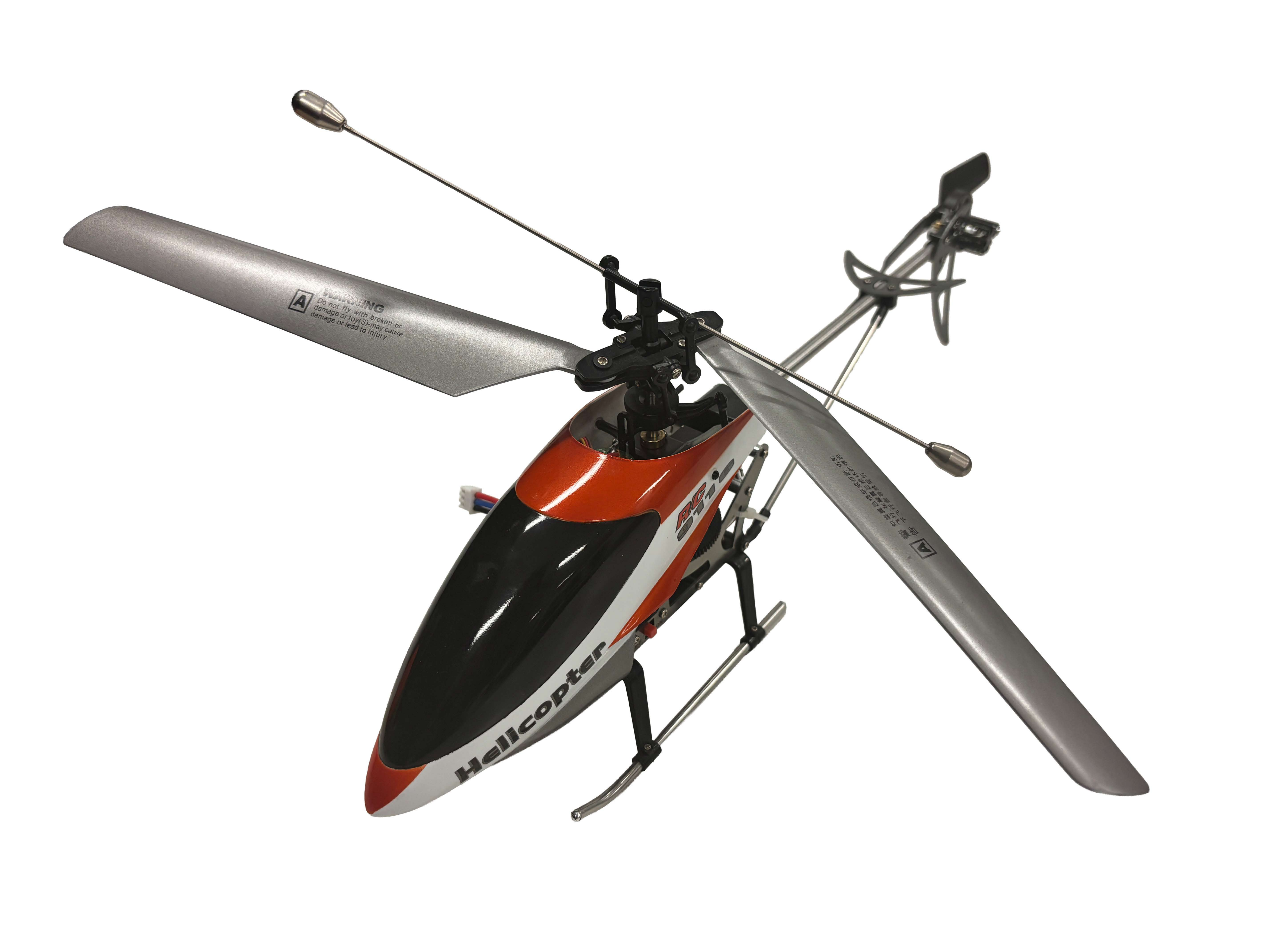}
\end{tabular}
& \small 14$\times$8$\times$39\,cm
& \small 234.96\,g
& \small Plastic \& Metal
& \small \texttt{helicopter}
\\

\begin{tabular}{@{}c@{}}
\includegraphics[width=0.05\textwidth]{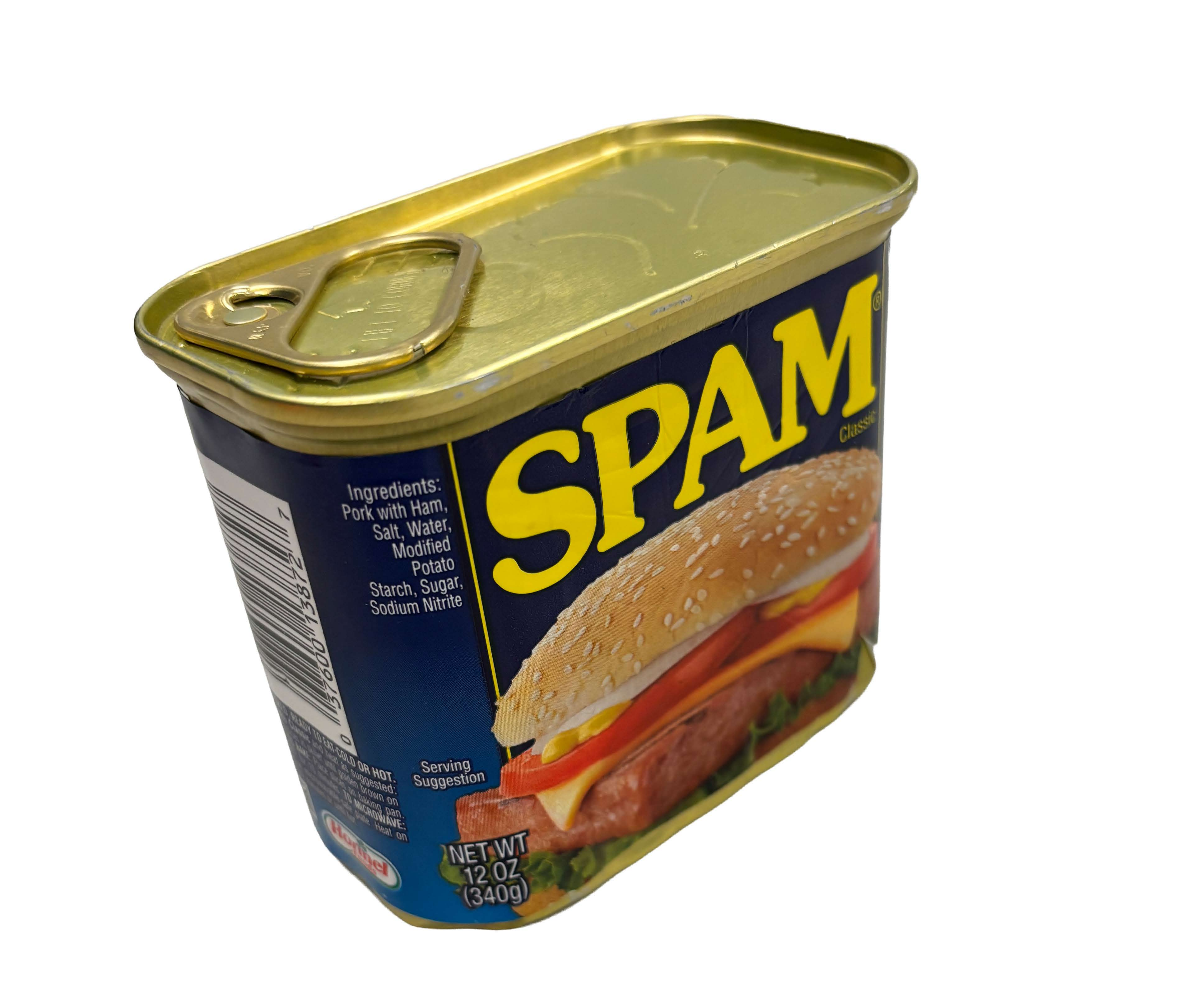}
\end{tabular}
& \small 8$\times$5.6$\times$10\,cm
& \small 367.41\,g
& \small Metal \& Spam
& \small \texttt{spam}
\\

\begin{tabular}{@{}c@{}}
\includegraphics[width=0.05\textwidth]{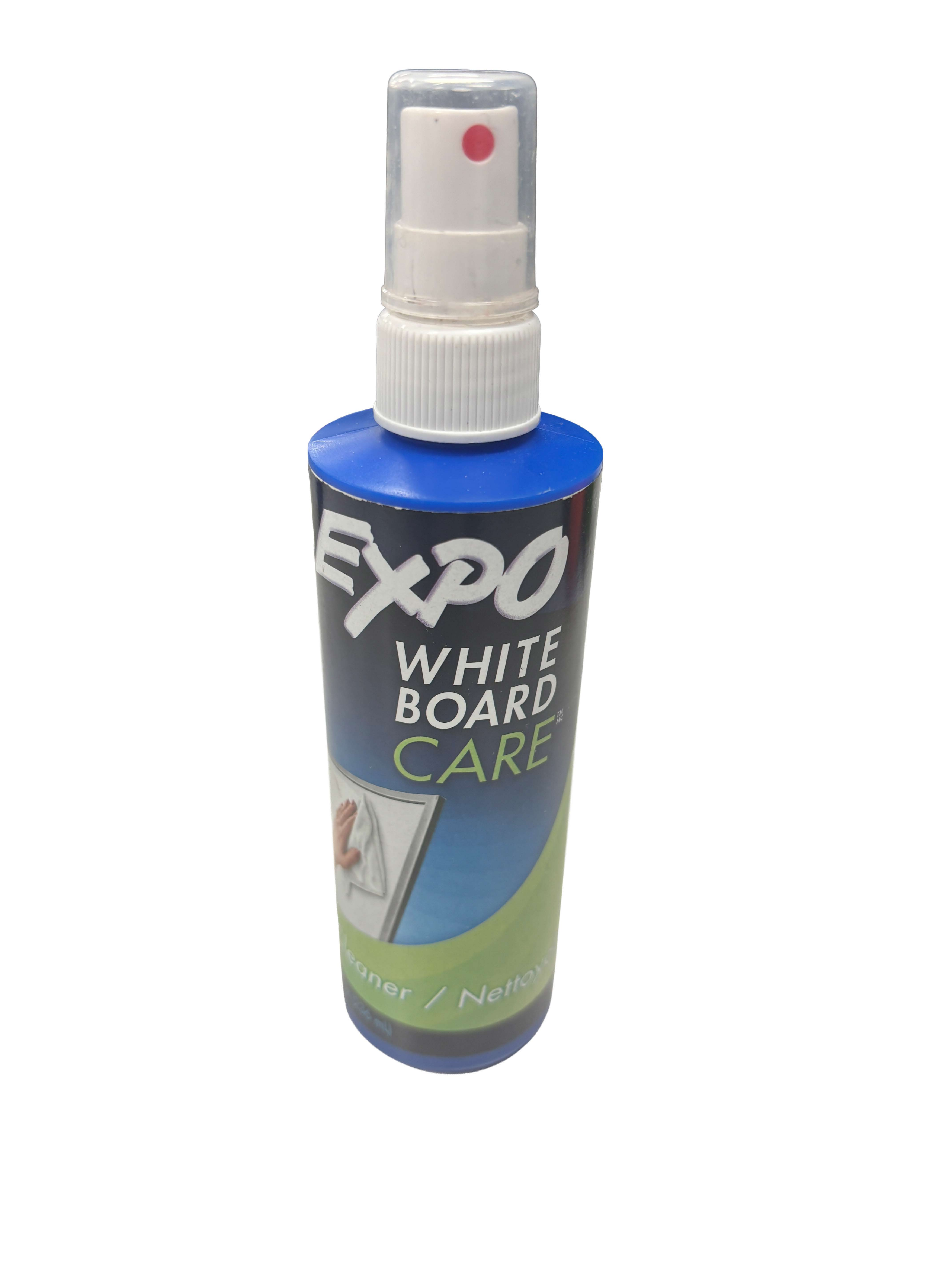}
\end{tabular}
& \small 19$\times$5$\times$5\,cm
& \small 86.18\,g
& \small Plastic \& Liquid
& \small \texttt{cleaner bottle}
\\

\midrule
\multicolumn{5}{@{}l}{\textit{Additional objects.}}\\
\midrule

\begin{tabular}{@{}c@{}}
\includegraphics[width=0.05\textwidth]{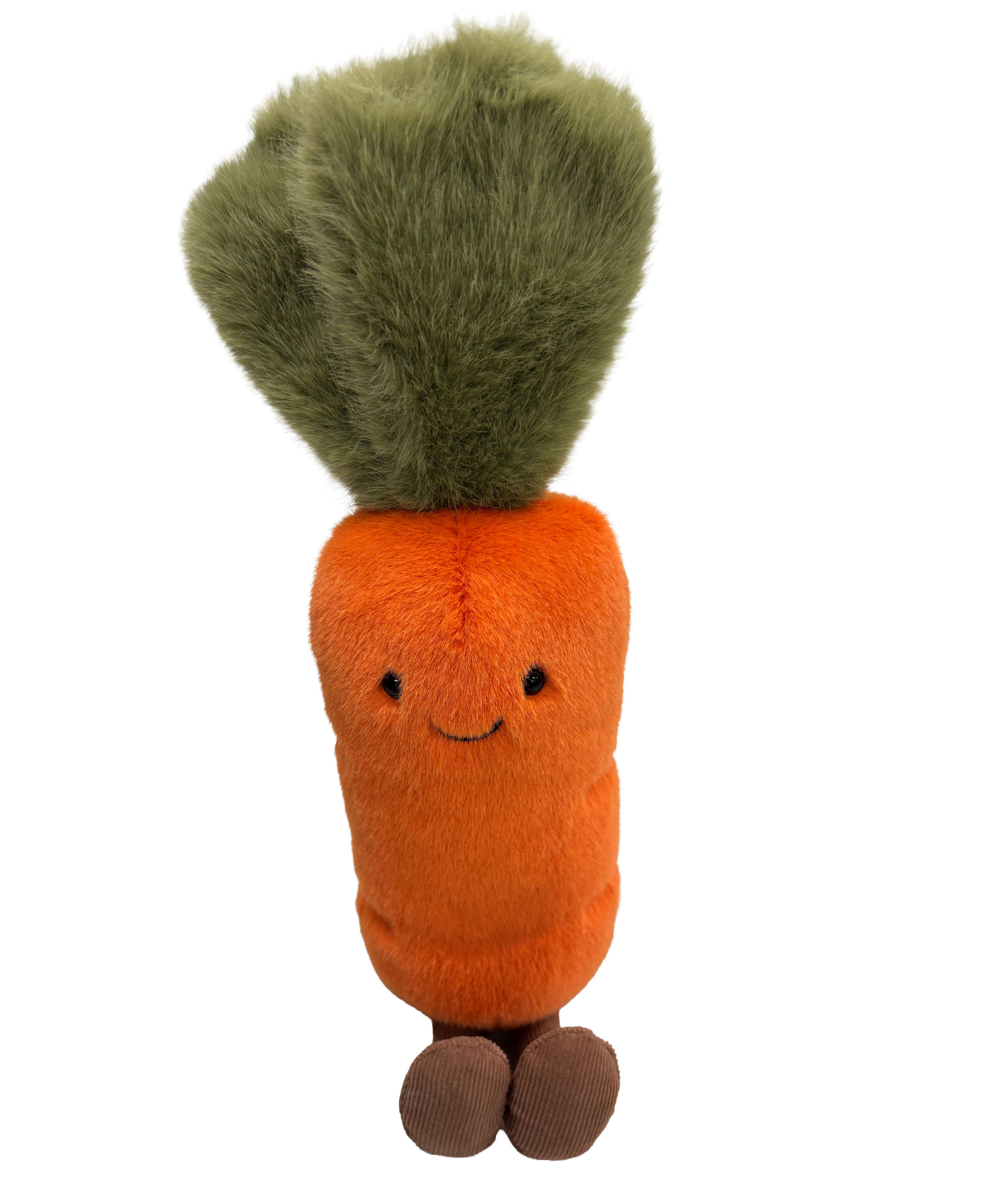}
\end{tabular}
& \small 24.5$\times$8$\times$6.5\,cm
& \small 135.17\,g
& \small Plush
& \small \texttt{carrot}
\\

\begin{tabular}{@{}c@{}}
\includegraphics[width=0.05\textwidth]{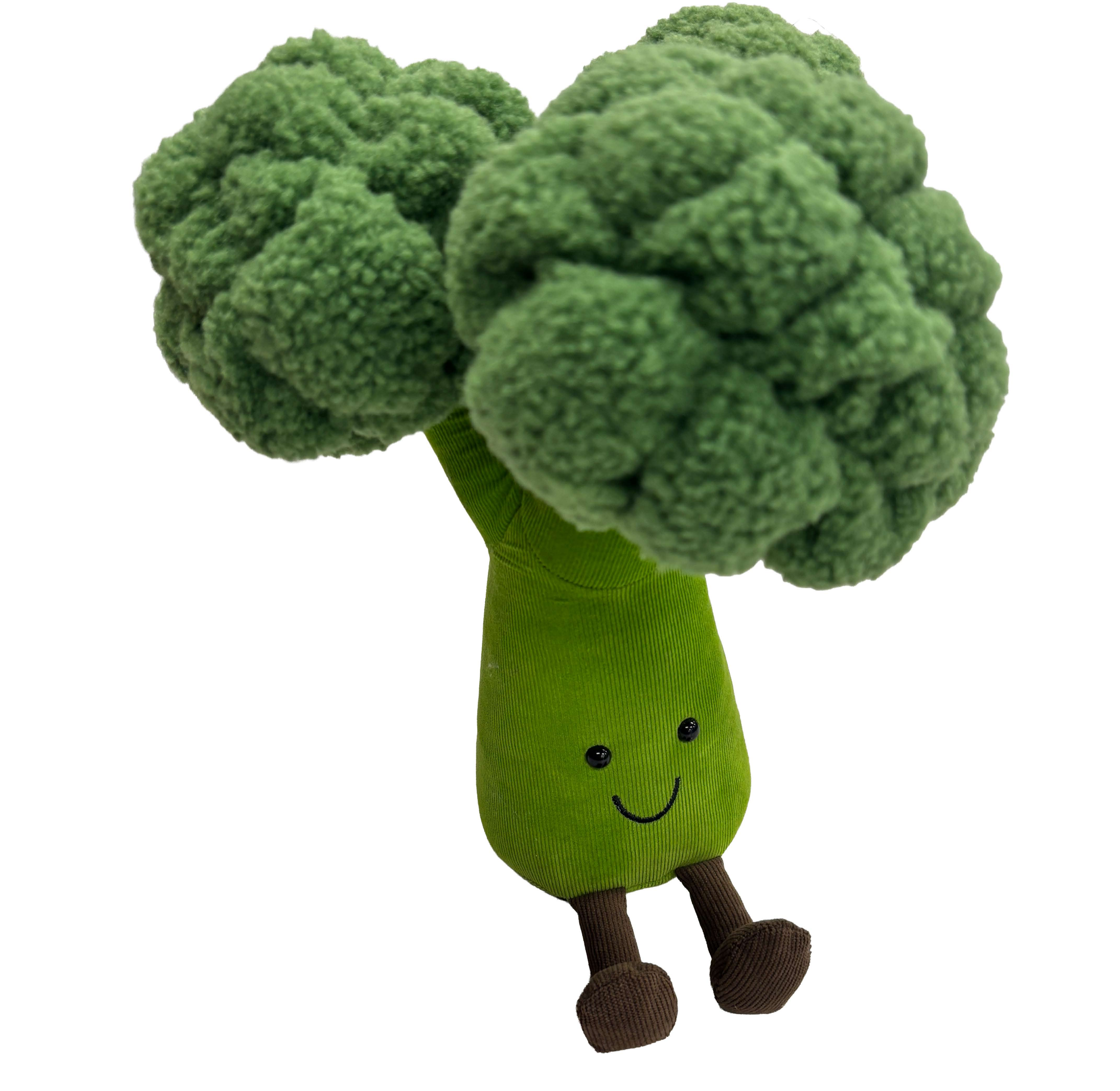}
\end{tabular}
& \small 23$\times$8$\times$8\,cm
& \small 307.08\,g
& \small  Plush 
& \small \texttt{broccoli}
\\

\begin{tabular}{@{}c@{}}
\includegraphics[width=0.05\textwidth]{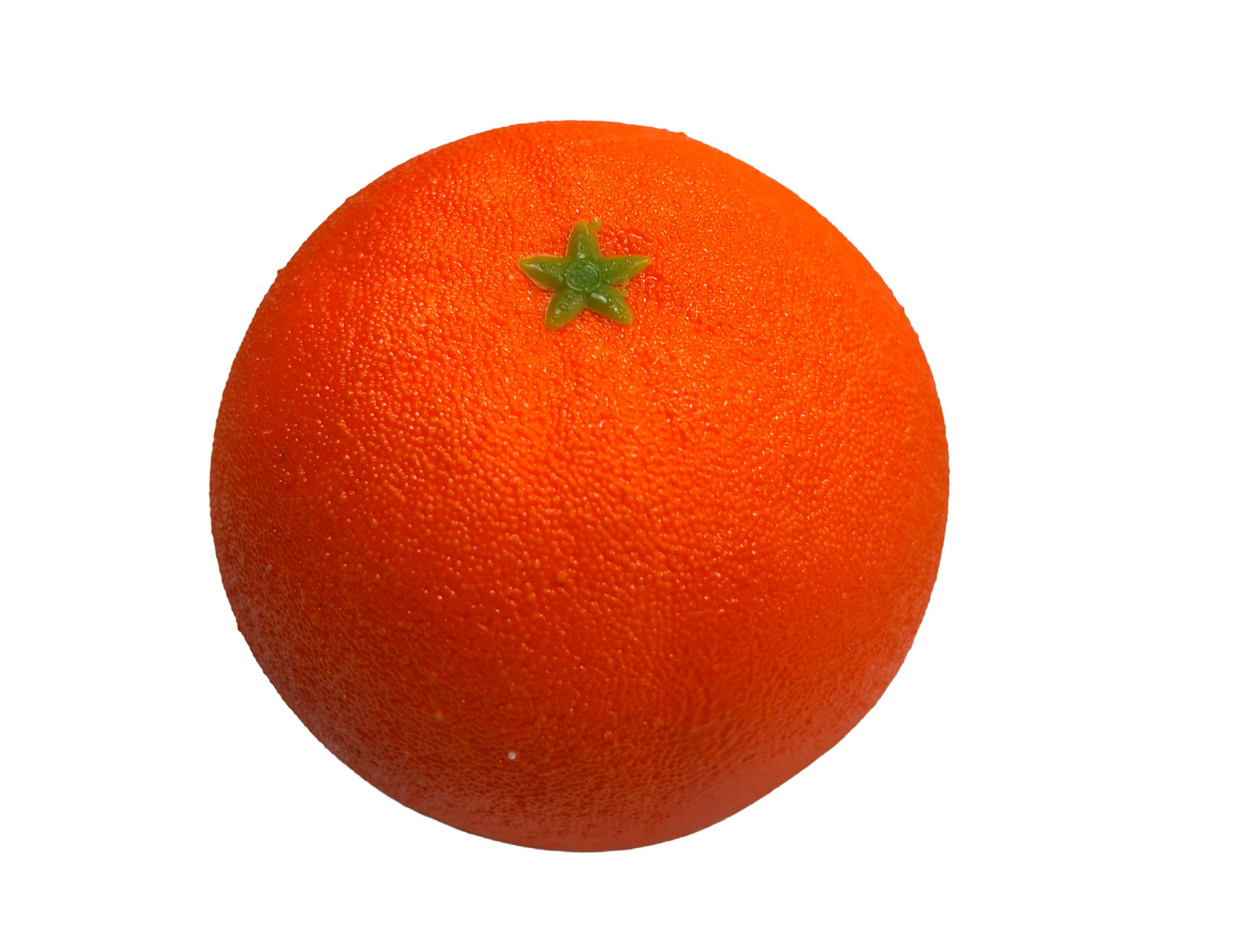}
\end{tabular}
& \small 7.5$\times$7.2$\times$7.5\,cm
& \small 19.05\,g
& \small  Plastic
& \small \texttt{orange}
\\

\begin{tabular}{@{}c@{}}
\includegraphics[width=0.05\textwidth]{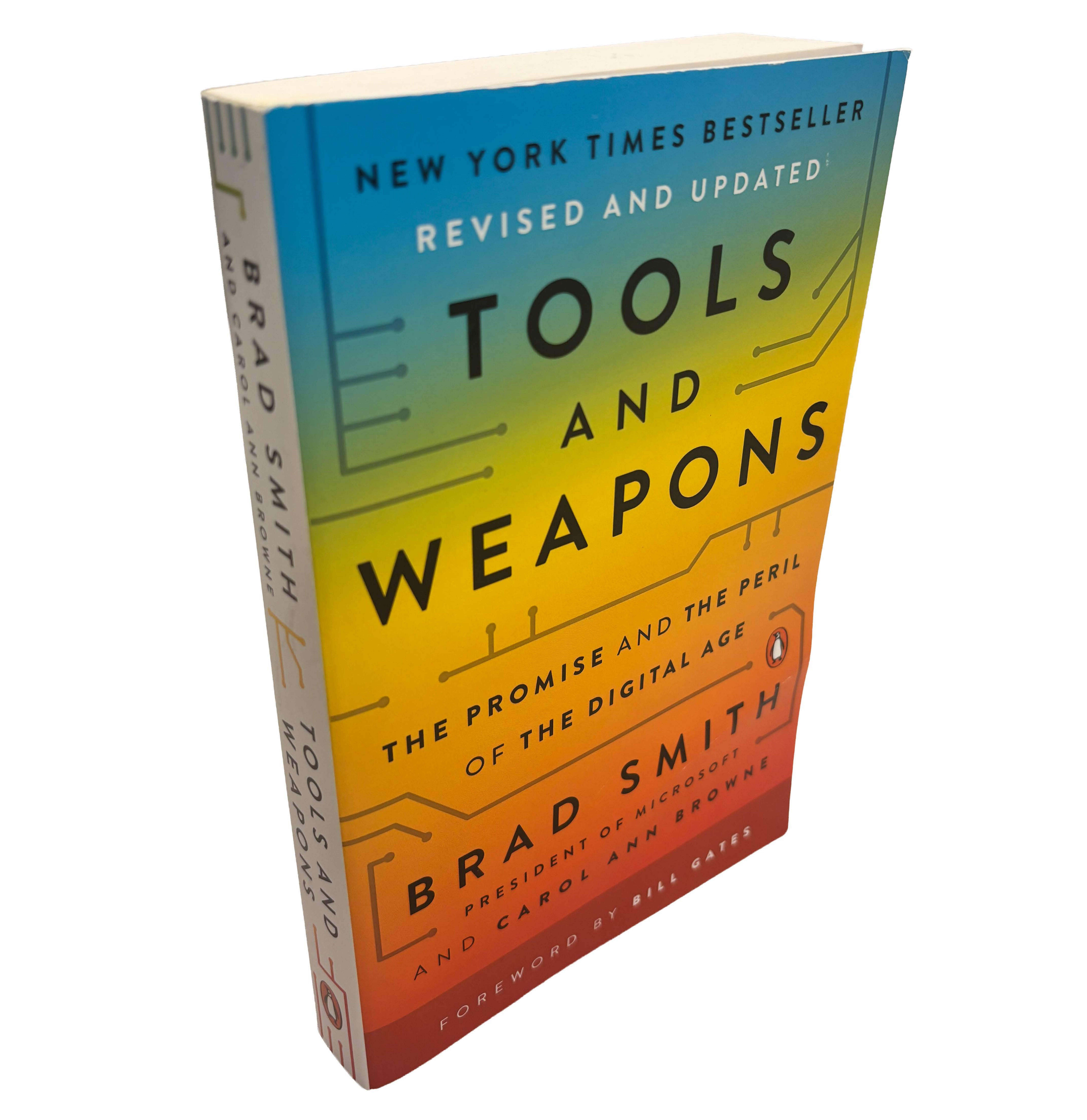}
\end{tabular}
& \small 21$\times$2.3$\times$13.8\,cm
& \small 376.03\,g
& \small Paper
& \small \texttt{book}
\\

\begin{tabular}{@{}c@{}}
\includegraphics[width=0.05\textwidth]{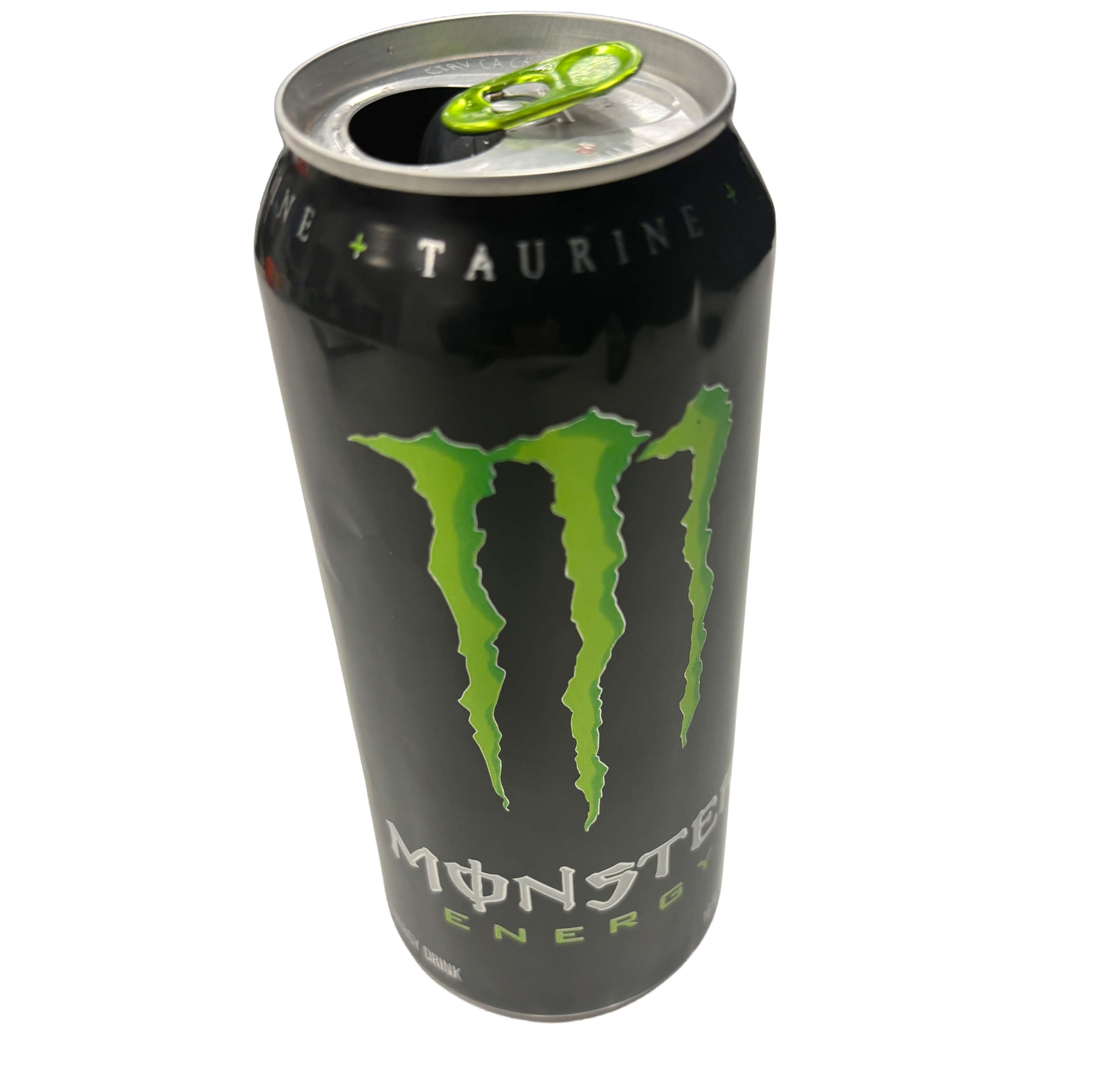}
\end{tabular}
& \small 15.3$\times$6.5$\times$6.5\,cm
& \small 18.14\,g
& \small Aluminum
& \small \texttt{black can}
\\

\end{xltabular}

\newcolumntype{Y}{>{\raggedright\arraybackslash}X}
\newcolumntype{C}[1]{>{\centering\arraybackslash}p{#1}}

\begin{table*}[!htbp]
\centering
\caption{\textbf{Testing scenes, table heights, language queries.}
Here we list the novel scenes chosen in this paper for evaluation, and the corresponding table height.
The snapshot of these scenes can be found in \cref{fig:benchmark_generalization}.
}
\label{tab:testing-scenes-details}

\setlength{\tabcolsep}{4pt}        % default is 6pt
\renewcommand{\arraystretch}{1.05} % default is 1.0

\begin{tabularx}{\linewidth}{@{} C{2.55cm} Y C{1.55cm} Y @{}}
\toprule
\textbf{Scene} & \textbf{Location} & \textbf{Table Height} & \textbf{Language Query}\\
\midrule

corridor
& \small CSL Studio \@ UIUC
& \small 0.43m
& \small \texttt{kettle}
\\

office lounge
& \small CSL Studio \@ UIUC
& \small 0.48m
& \small \texttt{toy dog}
\\

building café
& \small Siebel CS Building \@ UIUC
& \small 0.72m
& \small \texttt{Starbucks coffee}
\\

office
& \small CSL Studio \@ UIUC
& \small 0.74m
& \small \texttt{orange mug}
\\

building lounge
& \small Siebel CS Building \@ UIUC
& \small 0.74m
& \small \texttt{water bottle}
\\

office kitchenette
& \small CSL Studio \@ UIUC
& \small 0.74m
& \small \texttt{green apple}
\\

building den
& \small Siebel CS Building (RM 3333) \@ UIUC
& \small 0.74m
& \small \texttt{purple book}
\\

robotics lab
& \small CSL Studio \@ UIUC
& \small 0.86m
& \small \texttt{helicopter}
\\

office kitchen
& \small CSL Studio \@ UIUC
& \small 0.87m
& \small \texttt{spam}
\\

classroom
& \small Siebel CS Building (RM 1302) \@ UIUC
& \small 0.92m
& \small \texttt{cleaner bottle}
\\

\bottomrule
\end{tabularx}
\end{table*}

\end{appendices}

% ----------------------------------

\end{document}